\documentclass{article}

\PassOptionsToPackage{numbers, compress}{natbib}

\usepackage[final]{neurips_2025}

\usepackage[utf8]{inputenc} 
\usepackage[T1]{fontenc}    
\usepackage{hyperref}       
\usepackage{url}            
\usepackage{booktabs}       
\usepackage{amsfonts}       
\usepackage{nicefrac}       
\usepackage{microtype}      
\usepackage{xcolor}         
\usepackage{amsmath}
\usepackage{graphicx}
\usepackage{todonotes}
\usepackage{multirow}
\usepackage{multicol}
\usepackage{pifont}
\usepackage{color, colortbl}
\usepackage{wrapfig}
\usepackage{subcaption}
\usepackage[export]{adjustbox}
\usepackage{listings}
\usepackage{caption}

\definecolor{ForestGreen}{rgb}{0.13, 0.55, 0.13}
\definecolor{Green}{rgb}{0.0, 0.5, 0.0}
\definecolor{Blue}{rgb}{0.25, 0.42, 0.88}
\definecolor{green(munsell)}{rgb}{0.0, 0.66, 0.47}
\definecolor{green(ryb)}{rgb}{0.4, 0.69, 0.2}
\definecolor{green(pigment)}{rgb}{0.0, 0.65, 0.31}
\definecolor{citecolor}{HTML}{0071bc}
\definecolor{GrayXMark}{gray}{0.7}
\definecolor{DifferenceColor}{HTML}{af3235}
\definecolor{HighlightColor}{gray}{0.9}
\definecolor{OracleTextColor}{gray}{0.55}
\definecolor{Cerulean}{HTML}{00a2e3}

\def\tabref#1{Tab.~\ref{#1}}
\def\Tabref#1{Tab.~\ref{#1}}
\def\figref#1{Fig.~\ref{#1}}
\def\Figref#1{Fig.~\ref{#1}}

\title{InstanceAssemble: Layout-Aware Image Generation via Instance Assembling Attention
}

\author{%
  \textbf{Qiang Xiang$^{1,2}$,\ Shuang Sun$^{2}$,\ Binglei Li$^{1,3}$,}\\
  \textbf{Dejia Song$^2$,\ Huaxia Li$^{2}$, Yibo Chen$^{2}$, \ Xu Tang$^{2}$, Yao Hu$^{2}$, \ Junping Zhang$^{1}$\thanks{Corresponding author.}} \\
  $^1$Shanghai Key Laboratory of Intelligent Information Processing,\\ College of Computer Science and Artificial Intelligence, Fudan University\\ ~~\quad\quad $^2$Xiaohongshu Inc. ~~\quad\quad $^3$Shanghai Innovation Institute \\
  \texttt{\small \{qxiang24, blli24\}@m.fudan.edu.cn, jpzhang@fudan.edu.cn,}\\
  \texttt{\small \{sunshuang1, dejiasong, lihuaxia, zhaohaibo, tangshen, xiahou\}@xiaohongshu.com}  \\
}

\begin{document}

\maketitle

\vspace{-20pt}
\begin{figure}[ht]
\centering
{\includegraphics[clip, trim=0cm 71cm 0cm 0cm, width=0.98\textwidth]{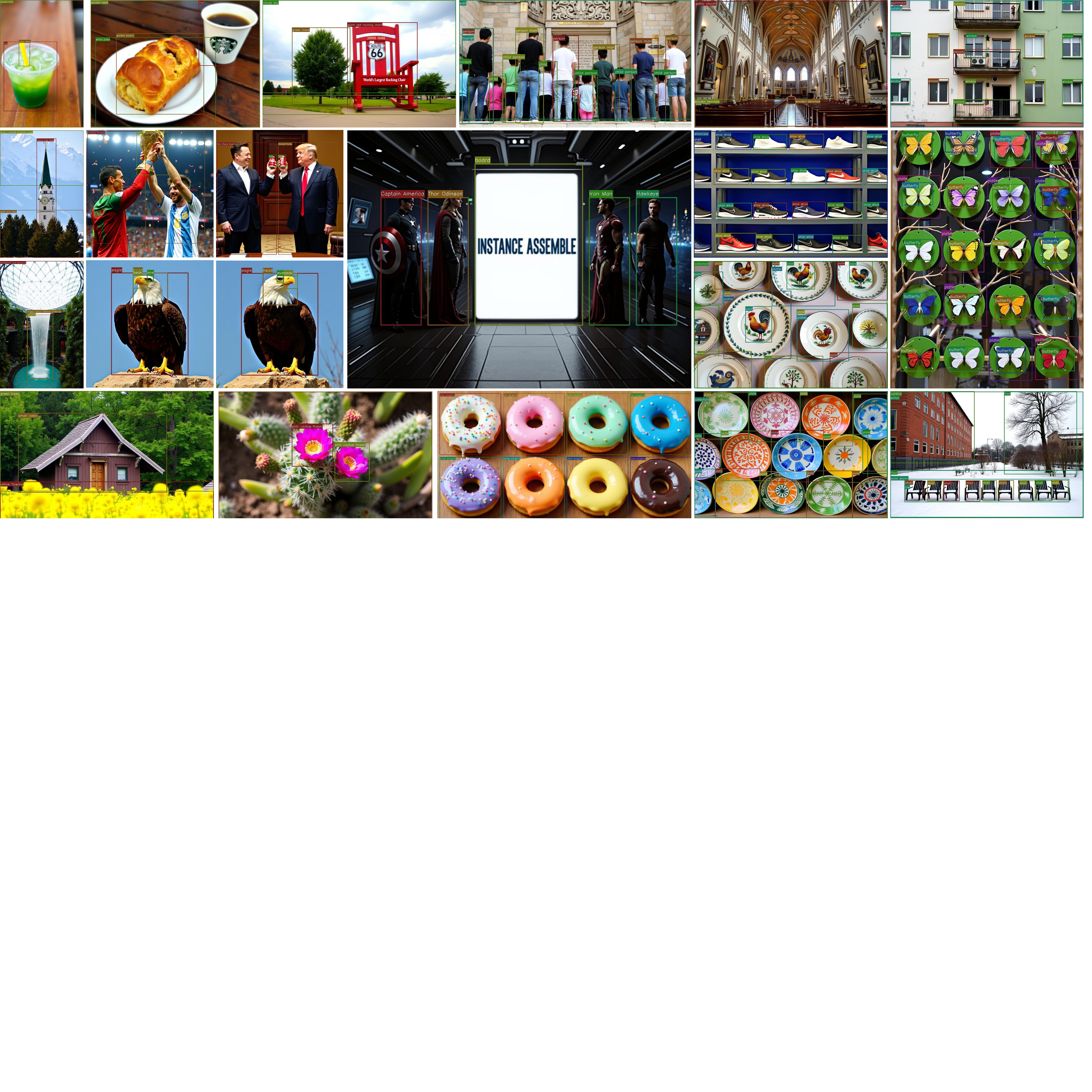}}
\caption{
\textbf{Layout-aware image generation result by InstanceAssemble}. We show image generation result under precise layout control, ranging from simple to intricate, sparse to dense layouts.}
\label{fig:teaser}
\end{figure}

\vspace{-5pt}
\begin{abstract}
\vspace{-5pt}
\label{sec:abstract}
Diffusion models have demonstrated remarkable capabilities in generating high-quality images. Recent advancements in Layout-to-Image (L2I) generation have leveraged positional conditions and textual descriptions to facilitate precise and controllable image synthesis. Despite overall progress, current L2I methods still exhibit suboptimal performance.
Therefore, we propose InstanceAssemble, a novel architecture that incorporates layout conditions via instance-assembling attention, enabling position control with bounding boxes (bbox) and multimodal content control including texts and additional visual content. Our method achieves flexible adaption to existing DiT-based T2I models through light-weighted LoRA modules. 
Additionally, we propose a Layout-to-Image benchmark, Denselayout, a comprehensive benchmark for layout-to-image generation, containing 5k images with 90k instances in total. We further introduce Layout Grounding Score (LGS), an interpretable evaluation metric to more precisely assess the accuracy of L2I generation. 
Experiments demonstrate that our InstanceAssemble method achieves state-of-the-art performance under complex layout conditions, while exhibiting strong compatibility with diverse style LoRA modules.
The code and pretrained models are publicly available at \url{https://github.com/FireRedTeam/InstanceAssemble}.
\end{abstract}
\section{Introduction}
\label{sec:intro}
Diffusion models~\cite{ddpm} have revolutionized image generation task, with architectures like Diffusion Transformer (DiT)~\cite{dit} offering superior quality over traditional UNet-based frameworks. Recent implementations such as Stable Diffusion 3/3.5~\cite{sd3, sd3.5} and Flux.1~\cite{flux} further enhance text-to-image alignment, paving the way for advancements in layout-controlled generation.
Layout-to-Image (L2I) generation is a task that focuses on creating images under layout conditions, allowing users to define spatial positions and semantic content of each instance explicitly. This task faces several significant challenges: (\textbf{\textit{i}}) ensuring precise layout alignment while maintaining high image quality, (\textbf{\textit{ii}}) preserving object positions and semantic attributes accurately during the iterative denoising process of diffusion models, and (\textbf{\textit{iii}}) supporting various types of reference conditions, such as texts, images and structure information. These challenges highlight the complexity of achieving robust and flexible layout-controlled image generation.

Existing L2I methods can be broadly categorized into training-free and training-based approaches, both possessing distinct advantages and limitations. 
Training-free methods~\cite{boxdiff, tflayoutguidance, beyourself, rag, regionalflux, groundit} rely on heuristic techniques without modifying the base model. However, these methods often exhibit degraded performance in complex layouts, demonstrate high sensitivity to hyperparameter tuning, and suffer from slow inference speed, which make them less practical for real-world applications. 
In contrast, training-based methods~\cite{layoutdiffusion, instancediffusion, migc, adversarial, creatilayout} involve training specific layout modules to improve layout alignment, which introduces a significant amount of extra parameters and increases training complexity and resource requirements.
Additionally, existing L2I evaluation metrics exhibit inaccuracies, such as false acceptance and localization errors. These identified shortcomings necessitate algorithm innovation for effective and efficient layout-controlled image generation.

Therefore, we propose \textbf{InstanceAssemble}, a novel framework that systematically tackles these issues through innovative design and efficient implementation. Our approach introduces a cascaded InstanceAssemble structure, which employs a multimodal interaction paradigm to process global prompts and instance-wise layout conditions sequentially. By leveraging the Assemble-MMDiT architecture, we apply an independent attention mechanism to the semantic content of each instance, thus enabling effective handling of dense and complex layouts. 
Furthermore, we adopt LoRA~\cite{lora} for lightweight adaptation, adding only 71M parameters to SD3-Medium (2B) and 102M to Flux.1 (11.8B). Our method enables position control with bounding boxes and multimodal content control including texts and additional visual content. This lightweight design preserves the capabilities of the base model while enhancing flexibility and efficiency.
We also introduce a novel metric called \textbf{Layout Grounding Score (LGS)} to ensure accurate evaluation for L2I generation, alongside a test dataset \textbf{DenseLayout}. This metric provides a consistent benchmark for assessing layout alignment. 

Our method achieves state-of-the-art performance across benchmarks and demonstrates robust layout alignment under a wide variety of scenarios, ranging from simple to intricate, sparse to dense layouts. 
Notably, despite being trained on sparse layouts ($\leq10$ instances), our approach maintains robust generalization capability on dense layouts ($\geq10$ instances), confirming the effectiveness of our proposed InstanceAssemble.
The main contributions are listed below.

\qquad  1. We propose a cascaded InstanceAssemble structure that processes global text prompts and layout conditions sequentially, enabling robust handling of complex layouts through an independent attention mechanism.

\qquad  2. By leveraging LoRA~\cite{lora}, we achieve efficient adaptation with minimal extra parameters (3.46\% on SD3-Medium and 0.84\% on Flux.1), supporting position control with multimodal content control while preserving capabilities of base model.

\qquad 3. We propose a new test dataset DenseLayout and a novel metric Layout Grounding Score (LGS) for Layout-to-Image evaluation. Experimental results demonstrate that our approach achieves state-of-the-art performance and robust capability under complex and dense layout conditions.

\section{Related Work}
\label{sec:related}
\textbf{Text-to-Image Generation}
Text-to-image synthesis~\cite{Imagen,ldm,sdxl,dit,hunyuan,lumina,pixart,allworthwords, dalle2, dalle3} has witnessed rapid progress with the development of diffusion models.
Initial works~\cite{dalle2, dalle3, Imagen, ldm} utilize UNet~\cite{unet} as the denoising architecture, leveraging cross-attention mechanisms to inject text-conditioning signals.
Recently, researches~\cite{pixart, sd3, sd3.5, flux, playgroundv3} have used the Multimodal Diffusion Transformer (MMDiT) architecture, marking a significant improvement.

\textbf{Layout-to-Image Generation}
Layout-to-Image generation enables image generation under layout conditions, which is defined as spatial positions with textual descriptions. Existing approaches can be broadly categorized into training-free and training-based paradigms.

\noindent\textbf{Training-free methods} leverage pretrained text-to-image diffusion models without additional training. A common strategy involves gradient-based guidance during denoising to align with layout conditions~\cite{boxdiff, zeroshotspatial, attnrefocus, beyourself, check, boxbind}. 
Also, there are methods that directly manipulate latents through well-defined replacing or merging operations~\cite{rag, noisecollage, multidiffusion} or enforce layout alignment via spatially constrained attention masks~\cite{regionalflux,he2023localizedtexttoimagegenerationfree}. 
GrounDiT~\cite{groundit} exploits semantic sharing in DiT: a cropped noisy patch and the full image become semantic clones when denoised together, enabling layout-to-image generation by jointly denoising instance regions with their corresponding image context.
Other approaches generate each instance separately and employ inpainting techniques to compose the final image~\cite{Zeropainter}.
However, these methods demonstrate decent performance primarily on simple and sparse layouts, while their accuracy decreases in more complex layouts. Some methods require hyperparameter tuning specific to different layout conditions, reducing their adaptability. Furthermore, additional gradient computations or latent manipulations result in slow inference speed, thus limiting their applicability in real-world scenarios.

\noindent\textbf{Training-based methods} explicitly incorporate layout conditioning through architectural modifications. Most approaches inject spatial constraints via cross-attention~\cite{layoutdiffusion, xue2023freestyle, place, ssmg, reco, ranni, ifadapter} or self-attention~\cite{layoutdiffuse, instancediffusion, gligen}. Some works propose dedicated layout encoding modules~\cite{hico, migc, migc++, lawdiff,controlnet} or adopt a two-stage pipeline that generates images after predicting a depth map with layout conditions~\cite{3dis,3disflux}. Other works leverage autoregressive image generation models~\cite{plangen}.
These methods suffer high computational costs due to excessive parameters.

\begin{figure}[t]
  \centering
  {\includegraphics[clip, trim=1cm 68cm 5cm 0cm, width=0.95\textwidth]{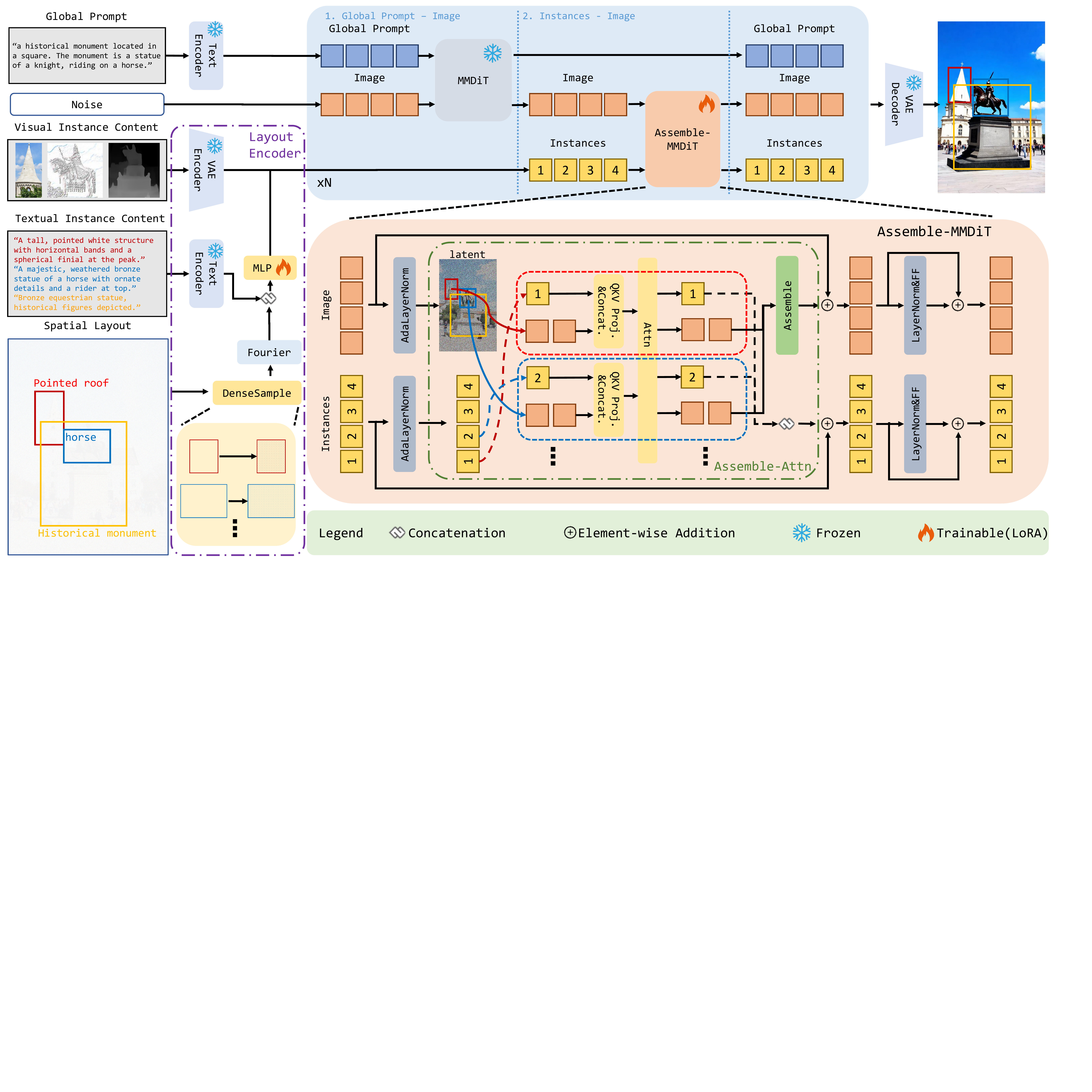}}
  \caption{\textbf{The proposed InstanceAssemble pipeline.} Various layout conditions are processed by the Layout Encoder to obtain instance tokens, which guide the image generation via Assemble-MMDiT. In Assemble-MMDiT, the instance tokens interact with image tokens through the Assembling-Attn.}
  \label{fig:arch}
\end{figure}

\section{Method}
\label{sec:method}
\textbf{Preliminaries}
Recent state-of-the-art text-to-image models such as SD3~\cite{sd3} and Flux~\cite{flux} adopt the Multimodal Diffusion Transformer (MMDiT) as the backbone for generation. 
Unlike traditional UNet-based cross-attention approaches, MMDiTs treat image and text modalities in a symmetric manner, which leads to stronger prompt alignment and controllability. 
These models are trained under the flow matching framework~\cite{flowmatching}, which formulates generation as learning a continuous velocity field that transports noise to data. 
Given a clean latent $\mathbf{x}$ and Gaussian noise $\boldsymbol{\epsilon} \sim \mathcal{N}(0,\mathbf{I})$, an interpolated latent is defined as
\begin{equation}
    \mathbf{z}_t = (1-t)\mathbf{x} + t\boldsymbol{\epsilon}, \quad t \in [0,1].
\end{equation}
The training objective minimizes the squared error between the predicted velocity and the target velocity $(\boldsymbol{\epsilon} - \mathbf{x})$:
\begin{equation}
    \mathcal{L}_{\text{FM}} = 
    \mathbb{E}_{\boldsymbol{\epsilon} \sim \mathcal{N}(0,\mathbf{I}), \,\mathbf{x}, \, t}
    \left[
    \left\|
    v_\theta(\mathbf{z}_t, t, \mathbf{y}) - (\boldsymbol{\epsilon} - \mathbf{x})
    \right\|_2^2
    \right],
\end{equation}
where $v_\theta$ is implemented with an MMDiT backbone.

\textbf{Problem Definition}
Layout-to-Image generation aims to synthesize images with precise control through a global prompt $\boldsymbol{p}$ and instance-wise layout conditions $\boldsymbol{L}$. The layout conditions comprise $N$ instances $\{l_i\}_{i=1}^N$, where each instance $l_i$ is defined by its spatial position $b_i$ and content $c_i$:
\begin{equation}
    \boldsymbol{L} = \{l_1, \dots, l_N\}, \quad \text{where} \ l_i = (c_i, b_i).
\end{equation}
In our framework, spatial positions are represented as bounding boxes, while instance content can be specified through multiple modalities: textual instance content and additional visual instance content, including reference images, depth maps and edge maps.

We propose InstanceAssemble, a framework with a \textbf{Layout Encoder} to encode the layout conditions and \textbf{Assemble-MMDiT} to effectively integrate the encoded layout conditions with image features.

\subsection{Layout Encoder}
We use a Layout Encoder~(\figref{fig:arch} left-bottom panel) to encode each instance $l_i$, and the tokens are denoted as $\boldsymbol{h}^{\boldsymbol{L}} = [h^{l_1}, \dots, h^{l_N}]$ which represents the layout information of each instance.
Given the spatial position of the instance (bounding box), we first enhance the spatial representation through \textbf{DenseSample}. Given a bounding box $b_i = (x_1, y_1, w, h) \in [0,1]^4$ with top-left coordinates $(x_1,y_1)$ and size $(w,h)$, we generate $K^2$ uniformly spaced points:
\begin{equation}
    \mathcal{P}_i = \left\{\left(x_1 + k_x \cdot \frac{w}{K}, y_1 + k_y \cdot \frac{h}{K}\right) \middle| k_x,k_y \in \{0,\dots,K-1\}\right\}
\end{equation}
Then, following GLIGEN\cite{gligen}, we compute the textual instance tokens as:
\begin{equation}
    h^i_l = \mathrm{MLP}\left([\boldsymbol{\tau}(c_i), \mathrm{Fourier}(\mathcal{P}_i)]\right),
\end{equation}
where $\boldsymbol{\tau}$ represents the text encoder, $\mathrm{Fourier}(\cdot)$ denotes Fourier embedding \cite{fourier}, [·, ·] denotes concatenation along the feature dimension, and MLP is a multi-layer perception.

Additionally, we can use additional visual instance content to better improve performance. Given the visual instance content, we first extract features using the VAE encoder of the base model, then project them to the unified instance token space through a MLP:
\begin{equation}
  h^i_l = \mathrm{MLP}\left(\mathrm{VAE}(c_i)\right).
\end{equation}
\subsection{Assemble-MMDiT}
We observe that applying attention between all image tokens and instance tokens results in suboptimal performance under complex layout conditions (e.g., overlapping, tiny objects). To address this, we introduce \textbf{Assemble-MMDiT} (\figref{fig:arch}, right-bottom panel), which enhances the location of each instance while maintaining compositional coherence with other instances.
Our method processes each instance independently through attention modules with its associated image tokens, followed by weighted feature assembling. 

Formally, given image tokens $\boldsymbol{h} \in \mathbb{R}^{C\times W\times H}$ (where $C$ denotes the latent channel size and $[W,H]$ the latent size) and instance tokens $\boldsymbol{h}^l \in \mathbb{R}^{C\times N}$, we apply AdaLayerNorm~\cite{adanorm}, followed by our proposed \textbf{Assembling-Attn}, as shown in \figref{fig:arch} (right-bottom panel).
We crop the image tokens $\boldsymbol{h}^z$ by the bbox $\boldsymbol{b}_i$ of each instance and get $\boldsymbol{h}^z_{l_i} = \boldsymbol{h}^z[\boldsymbol{b}_i] \in \mathbb{R}^{C\times w \times h}$. Then, we project the cropped image tokens $\boldsymbol{h}^z_{l_i}$ and their corresponding instance tokens $l_i$ into queries $(Q^{z_{l_i}}, Q^{l_i})$, keys $(K^{z_{l_i}}, K^{l_i})$, and values $(V^{z_{l_i}}, V^{l_i})$, and then apply attention:
\begin{equation} 
  \boldsymbol{h}^{z^\prime}_{l_i}, \boldsymbol{h}^{l_i^\prime} = \text{Attention}\left([Q^{z_{l_i}}, Q^{l_i}], [K^{z_{l_i}}, K^{l_i}], [V^{z_{l_i}}, V^{l_i}]\right). 
\end{equation}
where [·, ·] denotes concatenation along the token dimension. The updated tokens are assembled across instances. Let $M \in \mathbb{N}^{W\times H}$ represent the instance density map, calculating the counts of instances. The assembled image tokens $\boldsymbol{h}^{z^\prime}$ and instance tokens $\boldsymbol{h}^{l^\prime}$ are computed as:
\begin{equation} 
  \hspace{50pt}
  \begin{aligned} 
    \boldsymbol{h}^{z^\prime} :\boldsymbol{h}^{z^\prime}[:,i,j] &= \frac{1}{M[i,j]} \sum_{k=1}^N \boldsymbol{h}^{z^\prime}_{l_k}[:, i,j],\  \text{where} \ i\in [0,W-1], j\in[0, H-1]\\
    \boldsymbol{h}^{l^\prime} : \boldsymbol{h}^{l^\prime}[:,k]\ \ \,\,&=  \boldsymbol{h}^{l_k^\prime}. 
  \end{aligned} 
\end{equation}
\begin{wrapfigure}{r}{0.4\textwidth}
  \centering
  {\includegraphics[clip, trim=0cm 100cm 68cm 0cm,width=0.39\textwidth]{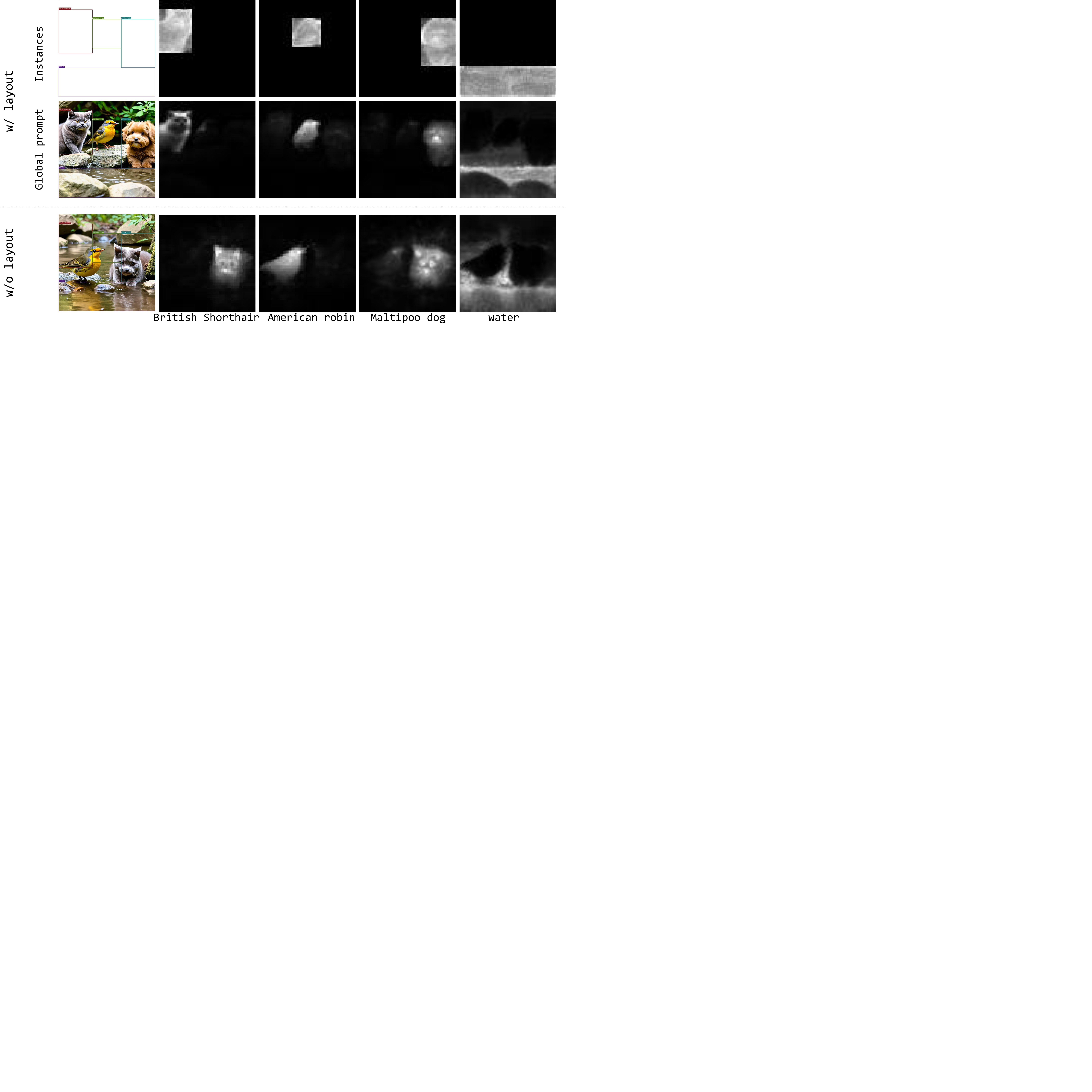}}
  \caption{(Top) instance-image attention map w/ layout. (Middle) global prompt-image attention map w/ layout. (Bottom) global prompt-image attention map w/o layout.} 
  \label{fig:attnmap}
\end{wrapfigure}
As illustrated in \figref{fig:attnmap}, the top row demonstrates that our assembling mechanism ensures instance tokens attend only to relevant image regions, where unrelated regions are left in black. The middle row reveals that the mechanism effectively guides global prompt tokens to focus on their correct spatial positions. In contrast, generation without explicit layout control (bottom row) results in localization errors ("British Shorthair" in wrong location) or semantic inconsistencies ("dog" missing).

Furthermore, to preserve the generation capability of the original model and mitigate conditional conflicts between global prompt and layout conditions, we employ a cascaded mechanism as shown in \figref{fig:arch} (right-above panel). In our design, the global text prompt and image latents are passed through original MMDiT first, then the image tokens along with instance tokens are processed by our Assemble-MMDiT module.
The first step captures global context and ensures generation quality, while the second step ensures instance layout alignment.
Besides, we train Assemble-MMDiT with LoRA, significantly reducing both the training cost and inference costs.

\begin{figure}[h]
  \centering
  {\includegraphics[clip, trim=0cm 112cm 15cm 0cm, width=0.8\textwidth]{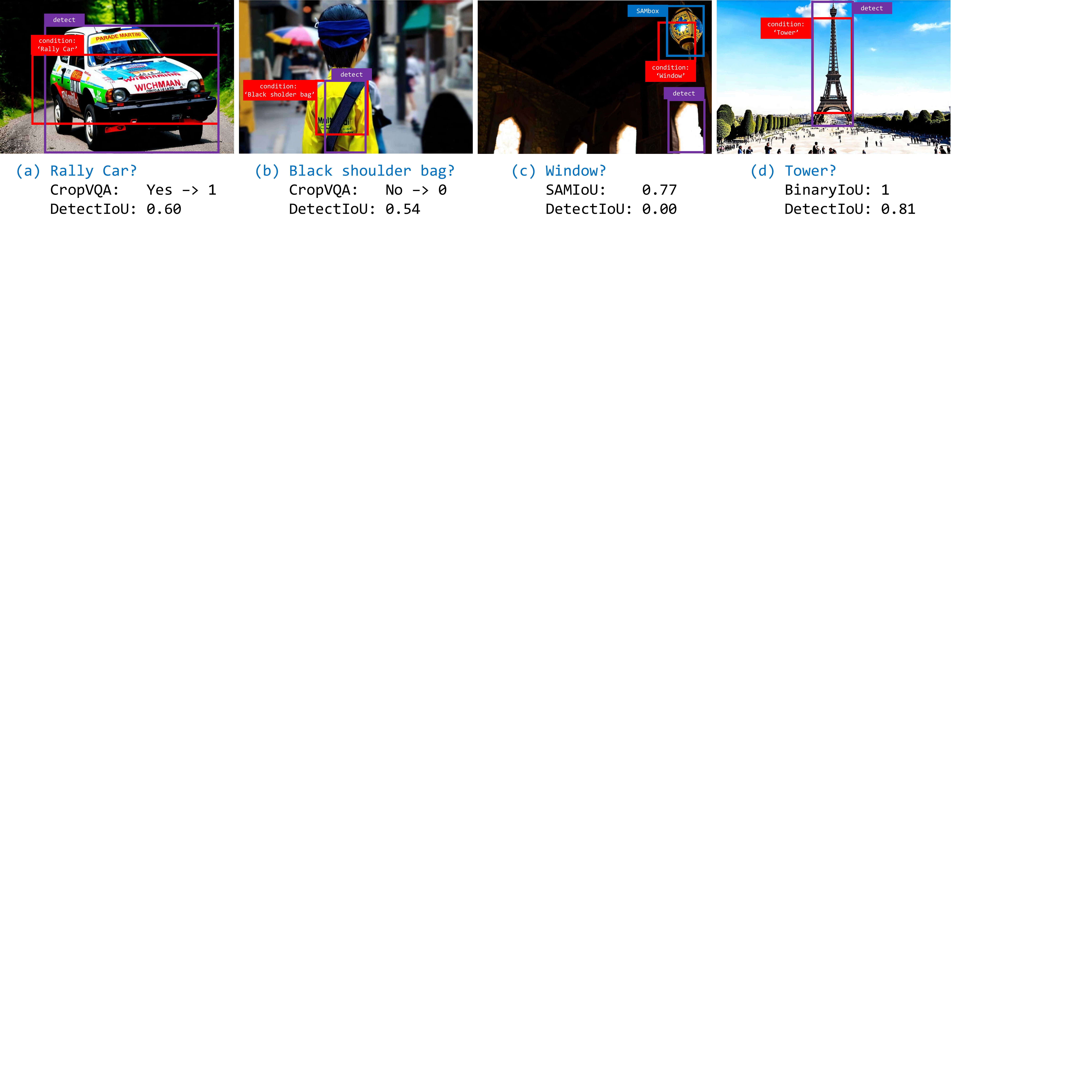}}
  \caption{\textbf{Failure cases of other metrics.} (\textbf{a}) false acceptance in CropVQA,(\textbf{b}) false rejection in CropVQA, (\textbf{c}) localization error in SAMIoU, and (\textbf{d}) discontinuous in BinaryIoU.} 
  \label{fig:metric}
\end{figure}
\subsection{Benchmark: DenseLayout and Layout Grounding Score}
The Layout-to-Image task aims to generate images that align precisely with provided layouts, evaluating both spatial accuracy and semantic consistency (e.g., color, texture, and shape, if provided).
The existing metrics (AP/AR) for object detection~\cite{layoutdiffuse,gligen,boxdiff,instancediffusion} are suboptimal. They assume a fixed category set and rely on inappropriate precision/recall for binary layout outcomes. VLM-based cropped VQA methods~\cite{creatilayout} suffer false acceptance (\figref{fig:metric} (a)) and false rejection (\figref{fig:metric} (b)). While spatial-only metrics like SAMIoU~\cite{rethink} ignore appearance consistency(\figref{fig:metric} (c)), GroundingDINO-based~\cite{groundingdino} binary IoU thresholds~\cite{migc,ifadapter} fail to capture continuous layout precision(\figref{fig:metric} (d)).
Thus, we propose \textbf{Layout Grounding Score (LGS)}, which integrates both spatial accuracy and semantic accuracy: 

\qquad 1. \textbf{Spatial Accuracy (DetectIoU)}: we detect all instances via an off-the-shelf detector~\cite{groundingdino}, compute the IoU against condition bbox, and report the global mean IoU across all instances for equal weighting.

\qquad 2. \textbf{Semantic Accuracy}: for instances with IoU>0.5, we crop the predicted region and assess the semantic accuracy by its attribute consistency~(color, texture, shape) via VLM-based VQA~\cite{minicpm}.

LGS supports open-set evaluation, uses DetectIoU to evaluate spatial accuracy and decouples the spatial and semantic check to avoid CropVQA~\cite{creatilayout} failures (shown in \figref{fig:metric}).
Furthermore, we introduce DenseLayout, a dense evaluation dataset for L2I, which consists of 5k images with 90k instances (18.1 per image).
The images in DenseLayout are generated by Flux.1-Dev, tagged by RAM++~\cite{ramplus}, detected by GroundingDINO~\cite{groundingdino}, recaptioned by Qwen2.5-VL~\cite{qwen2.5-VL}, and filtered to retain those with $\geq$15 instances, thus providing dense layout conditions.

\subsection{Training and Inference}
During training, we freeze the parameters of the base model and only update the proposed Layout Encoder and Assemble-MMDiT module. We denote the adding parameters by ${\theta}^{\prime}$. 
The training objective is given by
\begin{equation}
\mathcal{L}
= \mathbb{E}_{\mathbf{\epsilon} \sim \mathcal{N}(0,\mathbf{I}), \,\mathbf{x}, \, t, \,\mathbf{p}, \boldsymbol{L}}
\left[
\left\|
v_{\{\theta,\theta'\}}\!\big(\mathbf{z}_t, t, \mathbf{p}, \boldsymbol{L}\big)
-(\mathbf{\epsilon} - \mathbf{x})
\right\|_2^2
\right],
\label{eq:training_loss}
\end{equation}
where $\mathbf{z}_t = (1-t)\mathbf{x} + t\mathbf{\epsilon}$.
During inference, layout-conditioned denoising is applied during the first 30\% of diffusion steps, as the layout primarily forms in early stages~\cite{gligen,migc}.

\newcommand{\tablayousamsd}{
\begin{table}[t]
\scriptsize
\centering
\caption{\textbf{Quantitative comparison between our SD3-based InstanceAssemble and other L2I methods on LayoutSAM-Eval.} $^\star$ The CropVQA score is proposed in Creatilayout~\cite{creatilayout} and the score of InstanceDiff, MIGC and CreatiLayout is borrowed from CreatiLayout~\cite{creatilayout}.}
\label{tab:layousamsd}
\begin{tabular}{l*{11}{p{12pt}}}
\toprule
\multirow{2}{*}{LayoutSAM-Eval} & \multicolumn{4}{c}{CropVQA$^\star$} & \multicolumn{4}{c}{Layout Grounding Score} & \multicolumn{3}{c}{Global Quality} \\
 & spatial$\uparrow$ & color$\uparrow$ & texture$\uparrow$ & shape$\uparrow$ & mIoU$\uparrow$ & color$\uparrow$ & texture$\uparrow$ & shape$\uparrow$ & VQA$\uparrow$ & Pick$\uparrow$ & CLIP$\uparrow$ \\
\midrule
\textcolor{OracleTextColor}{Real Images(Upper Bound)} & \textcolor{OracleTextColor}{98.95} & \textcolor{OracleTextColor}{98.45} & \textcolor{OracleTextColor}{98.90} & \textcolor{OracleTextColor}{98.80} & \textcolor{OracleTextColor}{88.85} & \textcolor{OracleTextColor}{88.07} & \textcolor{OracleTextColor}{88.71} & \textcolor{OracleTextColor}{88.62} &  &  & \\
\midrule
InstanceDiff (SD1.5) & 87.99 & 69.16 & 72.78 & 71.08 & \underline{78.16} & \underline{63.14} & \textbf{66.82} & \textbf{65.86} & 86.42 & 21.16 & 11.73 \\
MIGC (SD1.4) & 85.66& 66.97 & 71.24 & 69.06 & 62.87 & 50.70 & 52.99 & 51.77 & 88.97 & 20.69 & 12.56 \\
HICO (realisticVisionV51) & 90.92 & 69.82 & 73.25 & 71.69 & 70.68 & 53.16 & 55.71 & 54.61 & 86.53 & \underline{21.77} & 9.47 \\
\midrule
CreatiLayout (SD3-M) & \underline{92.67} & \underline{74.45}  & \underline{77.21} & \underline{75.93} & 45.82 & 38.44 & 39.68 & 39.24 & \underline{92.74} & 21.71 & \textbf{13.82} \\
InstanceAssemble(ours) (SD3-M) & \textbf{94.97} & \textbf{77.53} & \textbf{80.72} & \textbf{80.11} & \textbf{78.88} & \textbf{63.89} & \underline{66.27}  & \textbf{65.86}  & \textbf{93.12}  & \textbf{21.79}  & \underline{12.76}\\
\bottomrule
\end{tabular}
\end{table}
}

\newcommand{\tabcocosd}{
\begin{wraptable}{r}{0.52\textwidth}
\scriptsize
\centering
\caption{\textbf{Comparison between our SD3-based InstanceAssemble and other L2I methods on COCO-Layout.} Since COCO don"t have detailed description for each instance, we cannot evaluate the attribute accuracy and only report the spatial accuracy - mIoU.}
\label{tab:cocosd}
\begin{tabular}{l*{4}{p{12pt}}}
\toprule
\multirow{2}{*}{COCO-Layout} & LGS & \multicolumn{3}{c}{Global Quality} \\
 & mIoU$\uparrow$ & VQA$\uparrow$ & Pick$\uparrow$ & CLIP$\uparrow$ \\
\midrule
\textcolor{OracleTextColor}{Real Images(Upper Bound)} & \textcolor{OracleTextColor}{49.14}  &  &  &  \\
\midrule
InstanceDiff (SD1.5) & \textbf{30.39} & 75.77 & 20.75 & 24.41 \\
MIGC (SD1.4) & 27.36 & 70.32 & 20.20 & 23.58 \\
HICO (realisticVisionV51) & 18.88 & 50.61 & 20.38 & 20.72 \\
CreatiLayout (SD3-M) & 7.12 & \underline{87.79} & \underline{21.22} & \underline{25.59} \\
InstanceAssemble(ours) (SD3-M) & \underline{27.85} & \textbf{89.06} & \textbf{21.58} & \textbf{25.68} \\ 
\bottomrule
\end{tabular}
\end{wraptable}
}

\newcommand{\tabledense}{
\begin{table}[t]
\scriptsize
\centering
\caption{\textbf{Quantitative comparison between our InstanceAssemble and other L2I methods on DenseLayout.}}
\label{tab:dense}
\begin{tabular}{l*{7}{p{12pt}}}
\toprule
\multirow{2}{*}{DenseLayout} & \multicolumn{4}{c}{Layout Grounding Score} & \multicolumn{3}{c}{Global Quality} \\
 & mIoU$\uparrow$ & color$\uparrow$ & texture$\uparrow$ & shape$\uparrow$ & VQA$\uparrow$ & Pick$\uparrow$ & CLIP$\uparrow$ \\
\midrule
\textcolor{OracleTextColor}{Real Images(Upper Bound)} & \textcolor{OracleTextColor}{92.35} & \textcolor{OracleTextColor}{76.52} & \textcolor{OracleTextColor}{80.78} & \textcolor{OracleTextColor}{79.78} &  &  &  \\
\midrule
InstanceDiff (SD1.5) & \underline{47.31} & \underline{29.48} & \underline{33.36} & \underline{32.43} & 88.79 & 20.87 & 11.73 \\
MIGC (SD1.4) & 34.39 & 22.10 & 23.99 & 23.45 & 91.18 & 20.74 & \underline{12.81} \\
HICO (realisticVisionV51) & 22.42 & 10.52 & 11.69 & 11.46 & 74.42 & 20.51 & 8.16 \\
\midrule
CreatiLayout (SD3-Medium) & 15.54 & 11.69 & 12.34 & 12.17 & \underline{93.42} & \textbf{21.88} & \textbf{12.89} \\
\textbf{InstanceAssemble}(ours) (SD3-Medium) & \textbf{52.07}  & \textbf{33.77}  & \textbf{36.21} & \textbf{35.81} & \textbf{93.54} & \underline{21.68}& 12.58 \\
\midrule
Regional-Flux (Flux.1-Dev) & 14.06 & 11.34 & 11.91 & 11.84 & {92.94}  & \textbf{22.67}   & 10.66  \\
RAG (Flux.1-Dev) & {17.23} & {14.22} & {14.62} & {14.55}  & 92.16 & \underline{22.28} & {11.01} \\
\textbf{InstanceAssemble}(ours) (Flux.1-Dev) & \underline{43.42}  & \underline{27.60}  & \underline{29.50} & \underline{29.14} & \underline{93.36} & 21.98 & \textbf{11.38}  \\
\textbf{InstanceAssemble}(ours) (Flux.1-Schnell) & \textbf{45.33}  & \textbf{27.73}  & \textbf{30.06} & \textbf{29.62} & \textbf{93.52} & 21.72 & \underline{10.78}  \\
\bottomrule
\end{tabular}
\end{table}
}

\newcommand{\tabdensevisual}{
\begin{table}[hbtp]
\scriptsize
\centering
\caption{\textbf{Quantitative results of additional visual content on DenseLayout.}}
\label{tab:densevisual}
\begin{tabular}{l*{7}{p{12pt}}}
\toprule
\multirow{2}{*}{DenseLayout} & \multicolumn{4}{c}{Layout Grounding Score} & \multicolumn{3}{c}{Global Quality} \\
 & mIoU$\uparrow$ & color$\uparrow$ & texture$\uparrow$ & shape$\uparrow$ & VQA$\uparrow$ & Pick$\uparrow$ &  CLIP$\uparrow$ \\
\midrule
\textcolor{OracleTextColor}{Real Images(Upper Bound)} & \textcolor{OracleTextColor}{92.35} & \textcolor{OracleTextColor}{76.52} & \textcolor{OracleTextColor}{80.78} & \textcolor{OracleTextColor}{79.78} &  &  &  \\
\midrule
text & 43.72 & 26.57 & 28.56 & 28.39 & \textbf{93.37} & 21.63 & 12.45 \\
text+image & \textbf{55.29} & \textbf{42.15} & \textbf{44.50} & \textbf{44.24} & 91.66 & \textbf{22.05}  & 12.95 \\
text+depth & 49.64 & 28.25 & 31.82 & 31.62 & 92.83 & 21.28 & 13.25 \\
text+edge & 50.73 & 29.45 & 33.92 & 33.84 & 90.13 & 21.26 & \textbf{13.55}\\
\bottomrule
\end{tabular}
\end{table}
}

\newcommand{\tabspeed}{
\begin{table}[t]
\scriptsize
\centering
\caption{\textbf{Parameter addition and time efficiency under sparse and dense layout conditions.} We evaluated on 10\% of the LayoutSAM-Eval and DenseLayout datasets at 1024×1024 resolution. $^{\star}$ are optimized for 512×512 resolution, so their results are reported at this scale.}
\label{tab:speed}
\begin{tabular}{lrrr}
\toprule
 & \multirow{2}{*}{\begin{tabular}[c]{c@{}c@{}}Parameter Addition\\ (relative parameter addition(\%)) \end{tabular}} & \multicolumn{2}{c}{Time Efficiency(s)~(relative runtime increase(\%))} \\
 &  & Sparse Layout & Dense Layout \\
\midrule
InstanceDiff$^{\star}$(SD1.5) & 369M (43\%) & 14.37 (+771\%) & 44.81 (+2754\%) \\
MIGC(SD1.4) & {57M} (6.64\%) & 14.41 (+25.4\%) & 21.58 (+87.5\%) \\
HICO$^{\star}$(realisticVisionV51) & 361M(33.9\%) & \underline{4.11} (+92.9\%) & 9.93 (+320\%) \\
CreatiLayout(SD3-M) & 1.2B (64.0\%) & 4.37 (+14.4\%) & \underline{4.42} (\textbf{+14.8\%}) \\
Regional-Flux(Flux.1-Dev) & - & 15.29 (+113\%) & 37.47 (+418\%) \\
RAG(Flux.1-Dev) & - & 15.69 (+119\%) & 21.14 (+192\%) \\
\midrule
InstanceAssemble(ours)(SD3-M) & {71M} ({3.46\%}) & 7.19 (+88.2\%) & 13.38 (+248\%) \\
InstanceAssemble(ours)(Flux.1-Dev) & \textbf{102M (0.84\%)} & 8.21 (\underline{+14.3\%}) & 10.28 (+41.9\%) \\
InstanceAssemble(ours)(Flux.1-Schnell) & \textbf{102M (0.84\%)} & \textbf{1.41 (+8.46\%)} & \textbf{1.70} (\underline{+28.8\%}) \\
\bottomrule
\end{tabular}
\end{table}
}

\newcommand{\tabablation}{
\begin{table}
\scriptsize
\centering
\caption{\textbf{Ablation study on our proposed components on DenseLayout.} 
"Assemble" refers to the presence of the Assemble-Attn module (architectural design). 
"Cascaded" indicates the interaction order: (\ding{52}) means global prompt–image interaction followed by instance–image interaction (cascaded structure), while (\ding{56}) means both are applied in parallel. 
"LoRA" specifies the training strategy for the Assemble-MMDiT module: (\ding{52}) indicates training with LoRA, while (\ding{56}) indicates full fine-tuning. 
"DenseSample" denotes whether the DenseSample spatial encoding is used.}
\label{tab:ablation}
\begin{tabular}{cccc*{5}{p{12pt}}}
\toprule
\multirow{2}{*}{Assemble} & \multirow{2}{*}{Cascaded} & \multirow{2}{*}{LoRA} & \multirow{2}{*}{DenseSample} & \multicolumn{4}{c}{Layout Grounding Score} &  \\
 &  &  &  & mIoU$\uparrow$ & color$\uparrow$ & texture$\uparrow$ & shape$\uparrow$ & VQA$\uparrow$  \\
 \midrule
\ding{56} & \ding{56} & \ding{56} & \ding{56} & 11.69 & 9.16 & 9.68 & 9.56 & 93.75  \\
\ding{52} & \ding{56} & \ding{56} & \ding{56} & 43.98 & 24.19 & 26.95 & 26.75 & 84.57  \\
\ding{52} & \ding{52} & \ding{56} & \ding{56} & 45.96 & 29.61 & 31.50 & 31.09 & 92.71 \\
\ding{52} & \ding{52} & \ding{52} & \ding{56} & 51.28 & 32.68 & 34.94 & 34.58 & 93.33  \\
\ding{52} & \ding{52} & \ding{52} & \ding{52} & \textbf{52.07} & \textbf{33.77} & \textbf{36.21} & \textbf{35.81} & \textbf{93.54}  \\
\bottomrule
\end{tabular}
\end{table}
}
\tablayousamsd
\tabledense

\section{Experiments}
\label{sec:experiments}
\subsection{Experimental Setup}
\textbf{Implementation Details}
The textual-only InstanceAssemble is trained on SD3-Medium~\cite{sd3} and Flux.1-Dev~\cite{flux} and the version with additional visual instance content is only trained on SD3-Medium. 
We freeze the pretrained MMDiT backbone and only adapt the Layout Encoder and LoRA modules of Assemble-MMDiT. 
Assemble-MMDiT is initialized from pretrained weights, and LoRA with rank=4 is applied. 
In the SD3-based model all Assemble-MMDiT blocks are adapted, while in the Flux-based model we adapt eight blocks (seven double-blocks and one single block) due to resource constraints. 
During inference, the LoRA-based Assemble-MMDiT is activated for the first 30\% of denoising steps, while the global prompt--image phase uses the frozen backbone. 
This design yields 71M (SD3-M) and 102M (Flux.1-Dev) additional parameters for the textual-only setting; the variant with additional visual instance content (SD3-M) introduces 85M parameters. 
All models are trained on LayoutSAM~\cite{creatilayout} at 1024$\times$1024 with Prodigy, for 380K iterations (batch size 2) on SD3-M and 300K iterations (batch size 1) on Flux.1-Dev, using 8$\times$H800 GPUs (7 days for SD3-M; 5 days for Flux.1-Dev).

\textbf{Evaluation Dataset} 
We use LayoutSAM-Eval~\cite{creatilayout} to evaluate performance on fine-grained open-set sparse L2I dataset, containing 5k images and 19k instances in total (3.8 instances per image). 
To assess performance on fine-grained open-set dense L2I evaluation dataset, we use the proposed \textit{DenseLayout}, which consists of 5k images and 90k instances in total (18.1 instances per image). 
Following conventional practice, we also evaluate on coarse-grained close-set L2I evaluation dataset COCO~\cite{coco}. We combine COCO-Stuff and COCO-Instance annotations to create our COCO-Layout evaluation dataset, containing 5k images and 57k instances in total (11.5 instances per image).

\textbf{Evaluation Metric}
We evaluate the accuracy of L2I generation using our proposed LGS metric along with CropVQA proposed by CreatiLayout~\cite{creatilayout}, measuring spatial and semantic accuracy. 
We also employ multiple established metrics to measure overall image quality and global prompt alignment, including VQA Score~\cite{vqascore}, PickScore~\cite{pickscore} and CLIPScore~\cite{clipscore}.

\subsection{Evaluation on L2I with Textual-Only Content}
\label{sec:eval_text_only}

\textbf{Fine-Grained Open-Set Sparse L2I Generation}
\tabref{tab:layousamsd} presents the quantitative results of InstanceAssemble on LayoutSAM-Eval~\cite{creatilayout}, reporting results using our proposed LGS, CropVQA~\cite{creatilayout} and global quality metrics. Our proposed InstanceAssemble not only achieves SOTA in spatial and semantic accuracy of each instance, but also demonstrates superior global quality.

\tabcocosd
\textbf{Fine-Grained Open-Set Dense L2I Generation} 
\tabref{tab:dense} presents results on DenseLayout. 
With the same SD3-Medium backbone, InstanceAssemble significantly outperforms CreatiLayout (mIoU: 52.07 vs. 15.54) while maintaining comparable global quality. 
On Flux.1, it also yields large gains over Regional-Flux and RAG (e.g., mIoU: 43.42 vs. 17.23 for RAG), showing that our cascaded Assemble-Attn design generalizes well across backbones. 
Compared to earlier UNet-based approaches such as InstanceDiff and MIGC, our method achieves higher spatial and semantic accuracy (mIoU: 52.07 vs. 47.31) without sacrificing realism. 
Overall, InstanceAssemble establishes consistent improvements in layout alignment while ensuring high image quality under challenging dense layouts.

\tabspeed
\textbf{Coarse-Grained Closed-Set L2I Generation}
\tabref{tab:cocosd} presents the quantitative result of InstanceAssemble on COCO-Layout. Our proposed InstanceAssemble surpasses previous methods in overall image quality but lags slightly behind InstanceDiff~\cite{instancediffusion} in layout precision. We attribute this gap to: (i) InstanceDiff’s fine-grained COCO training data with per-entity attribute annotations, and (ii) its entity-wise generation strategy, which improves precision at significant computational cost (\tabref{tab:speed}).

\begin{figure}[t]
  \centering
  {\includegraphics[clip, trim=0cm 74cm 54cm 0cm,width=0.98\textwidth]{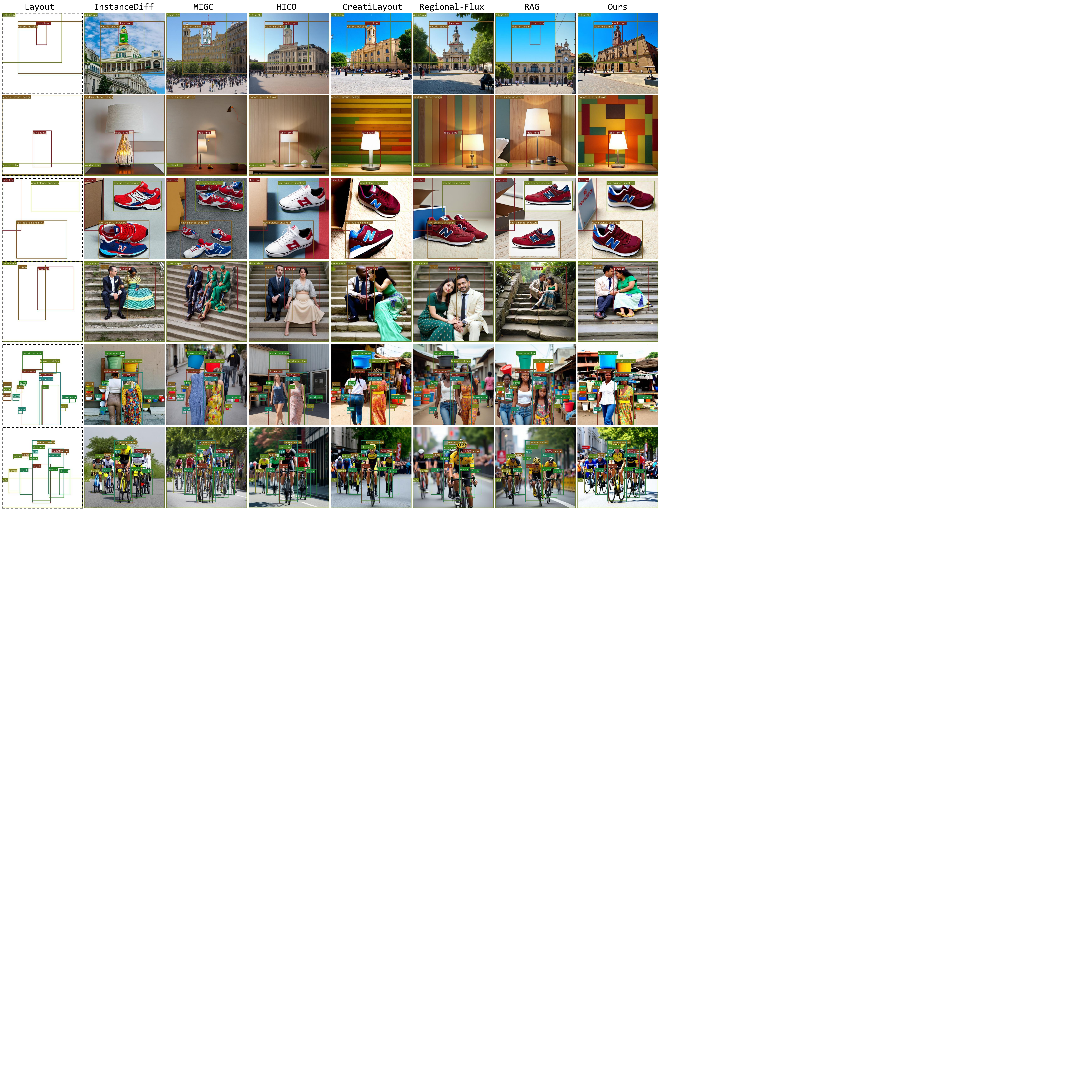}}
  \caption{\textbf{Qualitative comparison of InstanceAssemble with other methods.}}
  \label{fig:qualitative}
\end{figure}
\begin{wrapfigure}{r}{0.38\textwidth}
    \centering
    {\includegraphics[width=0.98\linewidth]{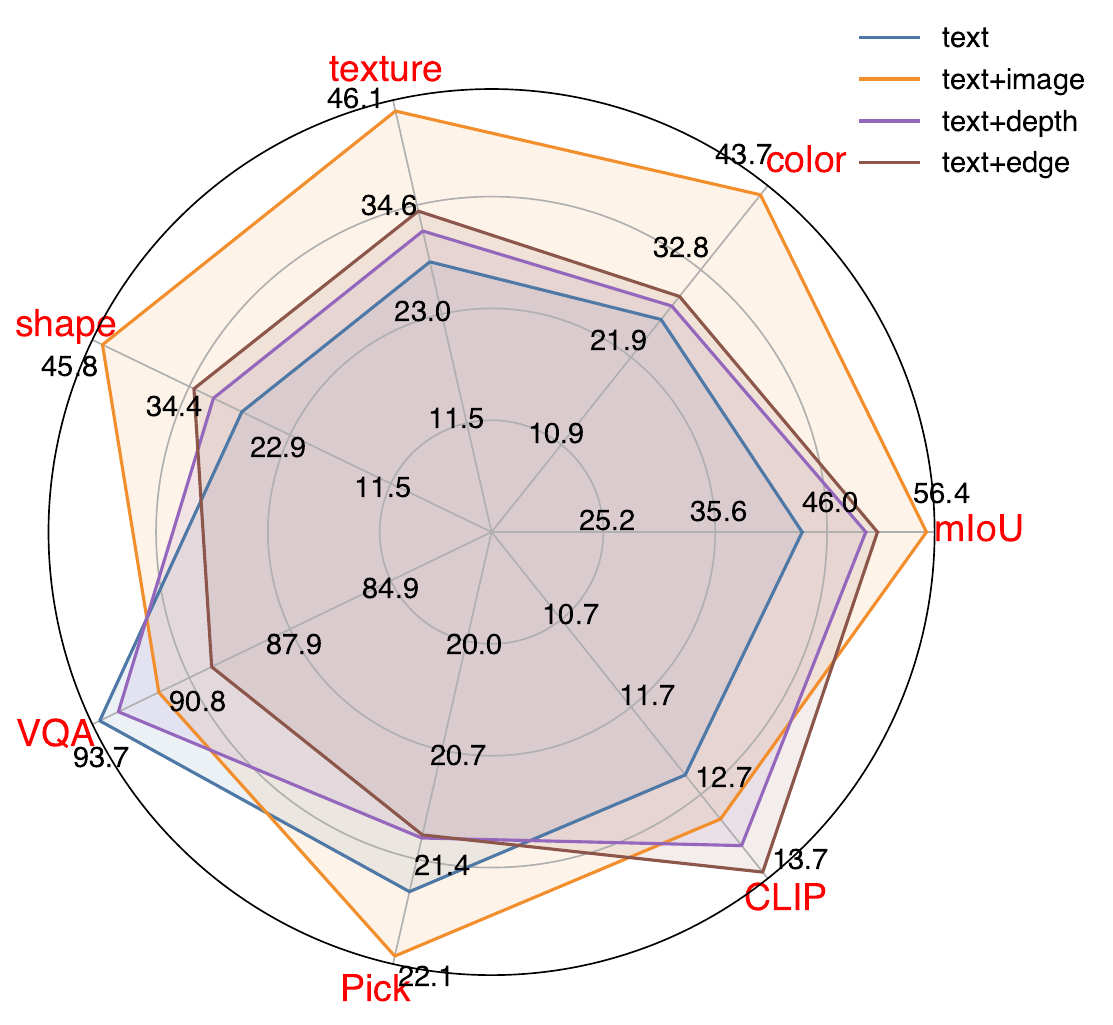}}
    \caption{\textbf{Quantitative results of InstanceAssemble with additional visual instance content.}} 
    \label{fig:visualradar}
\end{wrapfigure}
\textbf{Qualitative Comparison}
The comparative results in \figref{fig:qualitative} demonstrate that our proposed InstanceAssemble method achieves superior spatial precision and instance-caption alignment compared to baseline methods. For example, in the third row, both InstanceDiff~\cite{instancediffusion} and MIGC~\cite{migc} generate more than one shoes; HICO~\cite{hico} fails to generate the specified NewBalance shoe; Regional-Flux~\cite{regionalflux} does not adhere to the layout conditions; and the shoe generated by RAG~\cite{rag} is not properly fused with the background. In contrast, our method generates the correct instance, accurately placed and seamlessly integrated with the scene.

\textbf{Time Efficiency and Parameter Addition}
We compare time efficiency and parameter addition with other L2I methods, as shown in \tabref{tab:speed}. Our method achieves SOTA performance on layout alignment with acceptable time efficiency and minimal parameter addition.

\subsection{Evaluation on L2I with Additional Visual Content}
\label{sec:eval_addvis}
\begin{figure}[htbp]
    \centering
    {\includegraphics[clip, trim=0cm 93cm 17cm 0cm, width=0.98\textwidth]{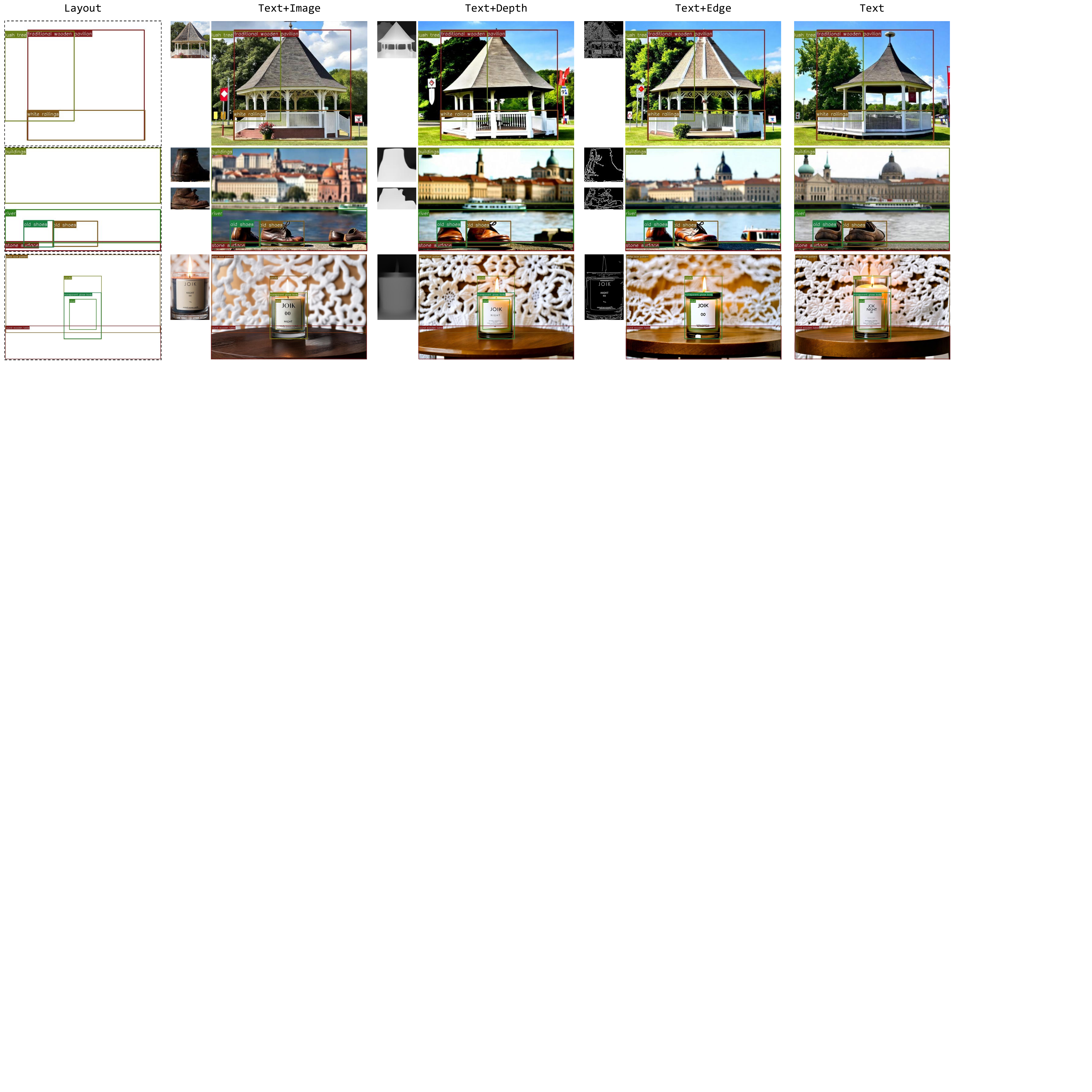}}
    \caption{\textbf{Qualitative results of InstanceAssemble with additional visual instance content.}} 
    \label{fig:visualqualitative}
\end{figure}

We evaluate three additional visual instance content: image, depth, and edge (see \figref{fig:visualradar}). Unsurprisingly, using image as additional instance content yields the best performance, as it provides rich visual information. Although depth and edge capture texture and shape features, their performance remains inferior compared to image instance content. Nevertheless, visual modalities outperform textual-only instance content. Qualitative comparisons (\figref{fig:visualqualitative}) further demonstrate that visual instance content leads to superior texture and shape alignment compared to text.

\tabablation
\subsection{Ablation Study}
We evaluated the contribution of each proposed component on SD3-M based InstanceAssemble in \tabref{tab:ablation}. 
The \textbf{base model} (SD3-Medium without additional modules) yields a very low Layout Grounding Score, indicating
poor layout and content control when instance information is not explicitly modeled.  
The introduction of Assemble-Attn module elevates spatial accuracy (mIoU to 43.98) and boosts semantic metrics (color/texture/shape to 24.19/26.95/26.75). 
The cascaded design (\ding{52}: prompt--image followed by instance--image; \ding{56}: parallel) resolves global quality
degradation while maintaining layout alignment.
Using \textbf{LoRA} to train Assemble-MMDiT improves performance for two reasons: 
(1) it retains the base model's capabilities compared to the fully fine-tuned version, and 
(2) it enables effective layout control with far fewer trainable parameters by introducing only lightweight low-rank matrices on attention projections. 
Finally, \textbf{DenseSample} further enhances spatial accuracy,
instance semantic accuracy and image quality. 
Together, these refinements progressively collectively optimize
layout to image modeling without compromising generation ability.

\subsection{Applications}
\begin{figure}[htbp]
  \centering
  {\includegraphics[clip, trim=3cm 85cm 7cm 4cm,width=0.98\textwidth]{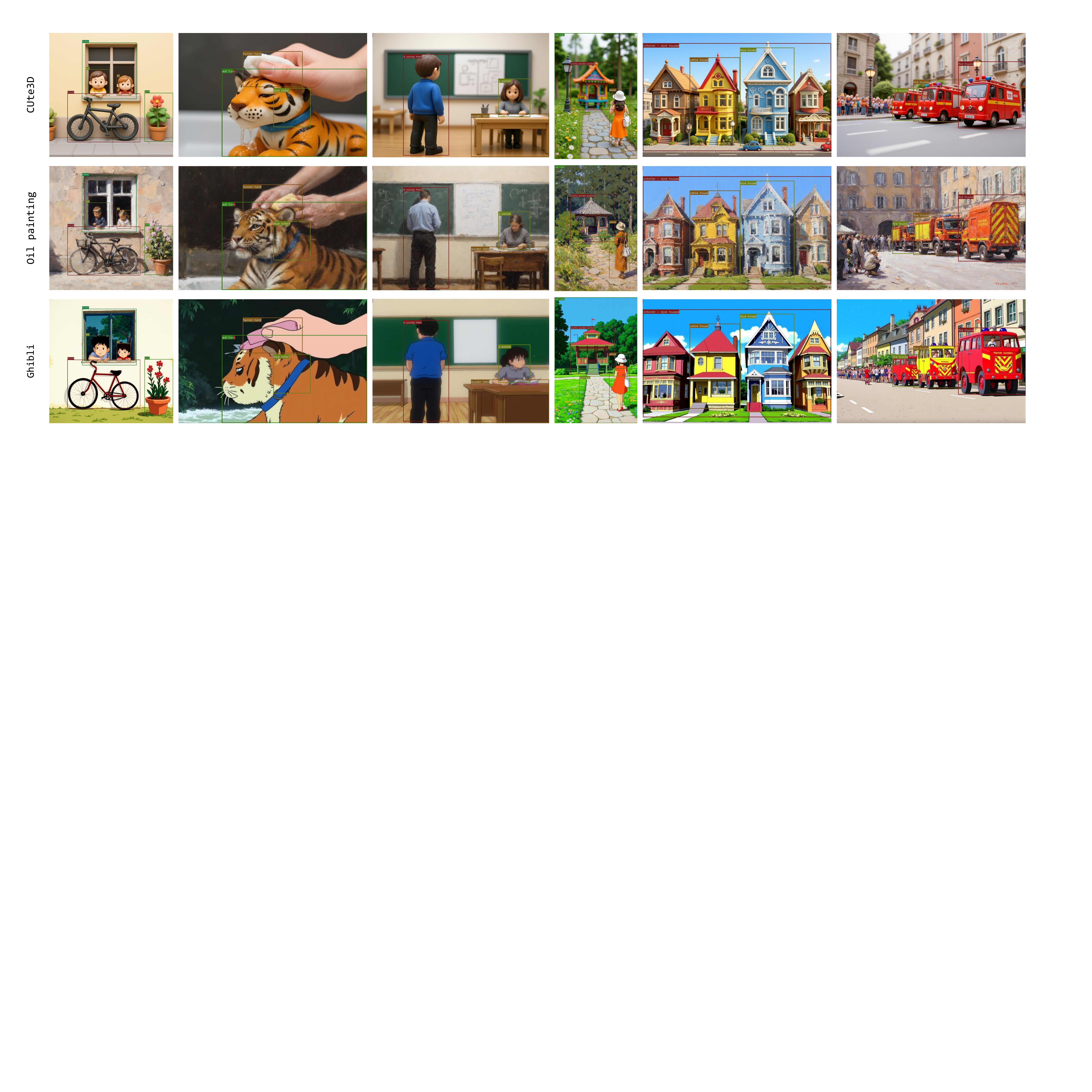}}
  \caption{\textbf{The adaption of Cute3D~\cite{cute3d}/ Oil Painting~\cite{oilpainting}/ Ghibli~\cite{ghibli} LoRA with our methods}. The proposed InstanceAssemble successfully adapts diverse style lora and maintaining superior layout alignment.}
  \label{fig:lora}
\end{figure}

We demonstrate that InstanceAssemble is versatile and applicable to various tasks. It seamlessly integrates with domain-specific LoRA modules for multi-domain style transfer while maintaining layout consistency, as shown in \figref{fig:lora}. 
Our proposed InstanceAssemble can cooperate with distilled models such as Flux.1-Schnell~\cite{flux}, as illustrated in \tabref{tab:dense}, achieving geometric layout control and detailed synthesis. Our approach demonstrates both style adaptability and computational efficiency, making it well-suited for controllable generative design applications.

\section{Conclusion}
\label{sec:conclusion}
We present InstanceAssemble, a novel approach for Layout-to-Image generation. Our method achieves state-of-the-art layout alignment while maintaining high-quality generation capabilities of DiT-based architectures.
We validate InstanceAssemble across textual instance content and additional visual instance content, demonstrating its versatility and robustness. Our layout control scheme also successfully adapts diverse style LoRAs while maintaining superior layout alignment, demonstrating cross-domain generalization capability. Futhermore, we introduce Layout Grounding Score metric and a DenseLayout evaluation dataset to validate performance under complex layout conditions.

\textbf{Limitations and Future Work} 
While our work advances controllable generation by unifying precise layout control with the expressive power of diffusion models, several limitations remain. 
First, our design currently requires sequential Assemble-MMDiT calls, which may incur inefficiency; exploring parallelization strategies is an important direction. 
Second, although our approach is effective under a wide range of layouts, image fidelity can degrade in extremely dense or highly complex cases.

\textbf{Broader Impacts} 
InstanceAssemble expands the frontier of structured visual synthesis by providing fine-grained layout control and high-quality multimodal generation. 
However, its powerful generative capabilities may also introduce risks. 
In particular, malicious use could enable the creation of misleading or deceptive layouts, exacerbating the spread of disinformation. 
The model may also raise privacy concerns if applied to sensitive data, and like many generative systems, it inherits and may amplify societal biases present in training corpora. 
We encourage responsible deployment and continued investigation into safeguards that mitigate these risks while enabling beneficial applications in design, education, and accessibility.

\newpage
\acksection
This research was supported by the National Natural Science Foundation of China (NSFC 62576103, 62176059). The computations were conducted using the CFFF platform at Fudan University. Part of this work was carried out during an internship at Xiaohongshu.
\bibliographystyle{abbrvnat}
\bibliography{main}

\begin{thebibliography}{66}
\providecommand{\natexlab}[1]{#1}
\providecommand{\url}[1]{\texttt{#1}}
\expandafter\ifx\csname urlstyle\endcsname\relax
  \providecommand{\doi}[1]{doi: #1}\else
  \providecommand{\doi}{doi: \begingroup \urlstyle{rm}\Url}\fi

\bibitem[Bao et~al.(2023)Bao, Nie, Xue, Cao, Li, Su, and Zhu]{allworthwords}
F.~Bao, S.~Nie, K.~Xue, Y.~Cao, C.~Li, H.~Su, and J.~Zhu.
\newblock {All are Worth Words: {A} ViT Backbone for Diffusion Models}.
\newblock In \emph{Proceedings of the IEEE/CVF Conference on Computer Vision and Pattern Recognition}, pages 22669--22679. {IEEE}, 2023.

\bibitem[Bar{-}Tal et~al.(2023)Bar{-}Tal, Yariv, Lipman, and Dekel]{multidiffusion}
O.~Bar{-}Tal, L.~Yariv, Y.~Lipman, and T.~Dekel.
\newblock {MultiDiffusion: Fusing Diffusion Paths for Controlled Image Generation}.
\newblock In \emph{Proceedings of the International Conference on Machine Learning}, volume 202 of \emph{Proceedings of Machine Learning Research}, pages 1737--1752. {PMLR}, 2023.

\bibitem[Betker et~al.(2023)Betker, Goh, Jing, Brooks, Wang, Li, Ouyang, Zhuang, Lee, Guo, et~al.]{dalle3}
J.~Betker, G.~Goh, L.~Jing, T.~Brooks, J.~Wang, L.~Li, L.~Ouyang, J.~Zhuang, J.~Lee, Y.~Guo, et~al.
\newblock {Improving image generation with better captions}, 2023.

\bibitem[{Black Forest Labs}(2024)]{flux}
{Black Forest Labs}.
\newblock Flux.
\newblock \url{https://github.com/black-forest-labs/flux}, 2024.

\bibitem[Chen et~al.(2024{\natexlab{a}})Chen, Xu, Zheng, Dai, Wang, Zhang, Wang, and Zhang]{regionalflux}
A.~Chen, J.~Xu, W.~Zheng, G.~Dai, Y.~Wang, R.~Zhang, H.~Wang, and S.~Zhang.
\newblock {Training-free Regional Prompting for Diffusion Transformers}, 2024{\natexlab{a}}.
\newblock URL \url{https://arxiv.org/abs/2411.02395}.

\bibitem[Chen et~al.(2024{\natexlab{b}})Chen, Yu, Ge, Yao, Xie, Wang, Kwok, Luo, Lu, and Li]{pixart}
J.~Chen, J.~Yu, C.~Ge, L.~Yao, E.~Xie, Z.~Wang, J.~T. Kwok, P.~Luo, H.~Lu, and Z.~Li.
\newblock {PixArt-{\(\alpha\)}: Fast Training of Diffusion Transformer for Photorealistic Text-to-Image Synthesis}.
\newblock In \emph{Proceedings of the International Conference on Learning Representations}. OpenReview.net, 2024{\natexlab{b}}.

\bibitem[Chen et~al.(2024{\natexlab{c}})Chen, Laina, and Vedaldi]{tflayoutguidance}
M.~Chen, I.~Laina, and A.~Vedaldi.
\newblock {Training-Free Layout Control with Cross-Attention Guidance}.
\newblock In \emph{Winter Conference on Applications of Computer Vision}, pages 5331--5341. {IEEE}, 2024{\natexlab{c}}.

\bibitem[Chen et~al.(2024{\natexlab{d}})Chen, Li, Wang, Chen, Jiang, Li, Wang, Yang, and Tai]{rag}
Z.~Chen, Y.~Li, H.~Wang, Z.~Chen, Z.~Jiang, J.~Li, Q.~Wang, J.~Yang, and Y.~Tai.
\newblock {Region-Aware Text-to-Image Generation via Hard Binding and Soft Refinement}, 2024{\natexlab{d}}.
\newblock URL \url{https://arxiv.org/abs/2411.06558}.

\bibitem[Cheng et~al.(2024{\natexlab{a}})Cheng, Ma, Wu, Liu, Ma, Wu, Leng, and Yin]{hico}
B.~Cheng, Y.~Ma, L.~Wu, S.~Liu, A.~Ma, X.~Wu, D.~Leng, and Y.~Yin.
\newblock {HiCo: Hierarchical Controllable Diffusion Model for Layout-to-image Generation}.
\newblock In \emph{Advances in Neural Information Processing Systems}, 2024{\natexlab{a}}.

\bibitem[Cheng et~al.(2023)Cheng, Liang, Shi, He, Xiao, and Li]{layoutdiffuse}
J.~Cheng, X.~Liang, X.~Shi, T.~He, T.~Xiao, and M.~Li.
\newblock {LayoutDiffuse: Adapting Foundational Diffusion Models for Layout-to-Image Generation}, 2023.
\newblock URL \url{https://arxiv.org/abs/2302.08908}.

\bibitem[Cheng et~al.(2024{\natexlab{b}})Cheng, Zhao, He, Xiao, Zhang, and Zhou]{rethink}
J.~Cheng, Z.~Zhao, T.~He, T.~Xiao, Z.~Zhang, and Y.~Zhou.
\newblock {Rethinking The Training And Evaluation of Rich-Context Layout-to-Image Generation}.
\newblock In \emph{Advances in Neural Information Processing Systems}, 2024{\natexlab{b}}.

\bibitem[Couairon et~al.(2023)Couairon, Careil, Cord, Lathuili{\`{e}}re, and Verbeek]{zeroshotspatial}
G.~Couairon, M.~Careil, M.~Cord, S.~Lathuili{\`{e}}re, and J.~Verbeek.
\newblock {Zero-shot spatial layout conditioning for text-to-image diffusion models}.
\newblock In \emph{Proceedings of the IEEE/CVF International Conference on Computer Vision}, pages 2174--2183. {IEEE}, 2023.

\bibitem[Dahary et~al.(2024)Dahary, Patashnik, Aberman, and Cohen{-}Or]{beyourself}
O.~Dahary, O.~Patashnik, K.~Aberman, and D.~Cohen{-}Or.
\newblock {Be Yourself: Bounded Attention for Multi-subject Text-to-Image Generation}.
\newblock In \emph{Proceedings of the European Conference on Computer Vision}, volume 15072 of \emph{Lecture Notes in Computer Science}, pages 432--448. Springer, 2024.

\bibitem[dewei Zhou et~al.(2025)dewei Zhou, Xie, Yang, and Yang]{3dis}
dewei Zhou, J.~Xie, Z.~Yang, and Y.~Yang.
\newblock {3{DIS}: Depth-Driven Decoupled Image Synthesis for Universal Multi-Instance Generation}.
\newblock In \emph{Proceedings of the International Conference on Learning Representations}, 2025.
\newblock URL \url{https://openreview.net/forum?id=MagmwodCAB}.

\bibitem[Esser et~al.(2024)Esser, Kulal, Blattmann, Entezari, M{\"{u}}ller, Saini, Levi, Lorenz, Sauer, Boesel, Podell, Dockhorn, English, and Rombach]{sd3}
P.~Esser, S.~Kulal, A.~Blattmann, R.~Entezari, J.~M{\"{u}}ller, H.~Saini, Y.~Levi, D.~Lorenz, A.~Sauer, F.~Boesel, D.~Podell, T.~Dockhorn, Z.~English, and R.~Rombach.
\newblock {Scaling Rectified Flow Transformers for High-Resolution Image Synthesis}.
\newblock In \emph{Proceedings of the International Conference on Machine Learning}. OpenReview.net, 2024.

\bibitem[Feng et~al.(2024)Feng, Gong, Chen, Shen, Liu, and Zhou]{ranni}
Y.~Feng, B.~Gong, D.~Chen, Y.~Shen, Y.~Liu, and J.~Zhou.
\newblock {Ranni: Taming Text-to-Image Diffusion for Accurate Instruction Following}.
\newblock In \emph{Proceedings of the IEEE/CVF Conference on Computer Vision and Pattern Recognition}, pages 4744--4753. {IEEE}, 2024.

\bibitem[Gao et~al.(2024)Gao, Zhuo, Liu, Du, Luo, Qiu, Zhang, Lin, Huang, Geng, Zhang, Xi, Shao, Jiang, Yang, Ye, Tong, He, Qiao, and Li]{lumina}
P.~Gao, L.~Zhuo, D.~Liu, R.~Du, X.~Luo, L.~Qiu, Y.~Zhang, C.~Lin, R.~Huang, S.~Geng, R.~Zhang, J.~Xi, W.~Shao, Z.~Jiang, T.~Yang, W.~Ye, H.~Tong, J.~He, Y.~Qiao, and H.~Li.
\newblock {Lumina-T2X: Transforming Text into Any Modality, Resolution, and Duration via Flow-based Large Diffusion Transformers}, 2024.
\newblock URL \url{https://arxiv.org/abs/2405.05945}.

\bibitem[Gong et~al.(2024)Gong, Huang, Feng, Zhang, Li, and Liu]{check}
B.~Gong, S.~Huang, Y.~Feng, S.~Zhang, Y.~Li, and Y.~Liu.
\newblock {Check, Locate, Rectify: {A} Training-Free Layout Calibration System for Text- to- Image Generation}.
\newblock In \emph{Proceedings of the IEEE/CVF Conference on Computer Vision and Pattern Recognition}, pages 6624--6634. {IEEE}, 2024.

\bibitem[He et~al.(2025)He, Cheng, Ma, Jia, Liu, Ma, Wu, Wu, Leng, and Yin]{plangen}
R.~He, B.~Cheng, Y.~Ma, Q.~Jia, S.~Liu, A.~Ma, X.~Wu, L.~Wu, D.~Leng, and Y.~Yin.
\newblock {PlanGen: Towards Unified Layout Planning and Image Generation in Auto-Regressive Vision Language Models}, 2025.
\newblock URL \url{https://arxiv.org/abs/2503.10127}.

\bibitem[He et~al.(2023)He, Salakhutdinov, and Kolter]{he2023localizedtexttoimagegenerationfree}
Y.~He, R.~Salakhutdinov, and J.~Z. Kolter.
\newblock Localized text-to-image generation for free via cross attention control, 2023.
\newblock URL \url{https://arxiv.org/abs/2306.14636}.

\bibitem[Hessel et~al.(2021)Hessel, Holtzman, Forbes, Bras, and Choi]{clipscore}
J.~Hessel, A.~Holtzman, M.~Forbes, R.~L. Bras, and Y.~Choi.
\newblock {CLIPScore: {A} Reference-free Evaluation Metric for Image Captioning}.
\newblock In \emph{Proceedings of the Empirical Methods in Natural Language Processing}, pages 7514--7528. Association for Computational Linguistics, 2021.

\bibitem[Ho et~al.(2020)Ho, Jain, and Abbeel]{ddpm}
J.~Ho, A.~Jain, and P.~Abbeel.
\newblock {Denoising Diffusion Probabilistic Models}.
\newblock In \emph{NeurIPS}, 2020.

\bibitem[Hu et~al.(2022)Hu, Shen, Wallis, Allen{-}Zhu, Li, Wang, Wang, and Chen]{lora}
E.~J. Hu, Y.~Shen, P.~Wallis, Z.~Allen{-}Zhu, Y.~Li, S.~Wang, L.~Wang, and W.~Chen.
\newblock {LoRA: Low-Rank Adaptation of Large Language Models}.
\newblock In \emph{Proceedings of the International Conference on Learning Representations}. OpenReview.net, 2022.

\bibitem[Huang et~al.(2023)Huang, Huang, Zhang, Tian, Feng, Zhang, Xie, Li, and Zhang]{ramplus}
X.~Huang, Y.-J. Huang, Y.~Zhang, W.~Tian, R.~Feng, Y.~Zhang, Y.~Xie, Y.~Li, and L.~Zhang.
\newblock {Open-Set Image Tagging with Multi-Grained Text Supervision}, 2023.
\newblock URL \url{https://arxiv.org/abs/2310.15200}.

\bibitem[Jia et~al.(2024)Jia, Luo, Dang, Dai, Chang, Wang, and Wang]{ssmg}
C.~Jia, M.~Luo, Z.~Dang, G.~Dai, X.~Chang, M.~Wang, and J.~Wang.
\newblock {{SSMG:} Spatial-Semantic Map Guided Diffusion Model for Free-Form Layout-to-Image Generation}.
\newblock In \emph{Proceedings of the AAAI Conference on Artificial Intelligence}, pages 2480--2488. {AAAI} Press, 2024.

\bibitem[Kirstain et~al.(2023)Kirstain, Polyak, Singer, Matiana, Penna, and Levy]{pickscore}
Y.~Kirstain, A.~Polyak, U.~Singer, S.~Matiana, J.~Penna, and O.~Levy.
\newblock {Pick-a-Pic: An Open Dataset of User Preferences for Text-to-Image Generation}.
\newblock In \emph{Advances in Neural Information Processing Systems}, 2023.

\bibitem[Lee et~al.(2024)Lee, Yoon, and Sung]{groundit}
Y.~Lee, T.~Yoon, and M.~Sung.
\newblock {GrounDiT: Grounding Diffusion Transformers via Noisy Patch Transplantation}.
\newblock In \emph{Advances in Neural Information Processing Systems}, 2024.

\bibitem[Li et~al.(2023)Li, Liu, Wu, Mu, Yang, Gao, Li, and Lee]{gligen}
Y.~Li, H.~Liu, Q.~Wu, F.~Mu, J.~Yang, J.~Gao, C.~Li, and Y.~J. Lee.
\newblock {{GLIGEN:} Open-Set Grounded Text-to-Image Generation}.
\newblock In \emph{Proceedings of the IEEE/CVF Conference on Computer Vision and Pattern Recognition}, pages 22511--22521. {IEEE}, 2023.

\bibitem[Li et~al.(2024{\natexlab{a}})Li, Keuper, Zhang, and Khoreva]{adversarial}
Y.~Li, M.~Keuper, D.~Zhang, and A.~Khoreva.
\newblock {Adversarial Supervision Makes Layout-to-Image Diffusion Models Thrive}.
\newblock In \emph{Proceedings of the International Conference on Learning Representations}. OpenReview.net, 2024{\natexlab{a}}.

\bibitem[Li et~al.(2024{\natexlab{b}})Li, Zhang, Lin, Xiong, Long, Deng, Zhang, Liu, Huang, Xiao, Chen, He, Li, Li, Zhang, Quan, Lu, Huang, Yuan, Zheng, Li, Zhang, Zhang, Chen, Liu, Fang, Wang, Xue, Tao, Zhu, Liu, Lin, Sun, Li, Wang, Chen, Hu, Xiao, Chen, Liu, Liu, Wang, Yang, Jiang, and Lu]{hunyuan}
Z.~Li, J.~Zhang, Q.~Lin, J.~Xiong, Y.~Long, X.~Deng, Y.~Zhang, X.~Liu, M.~Huang, Z.~Xiao, D.~Chen, J.~He, J.~Li, W.~Li, C.~Zhang, R.~Quan, J.~Lu, J.~Huang, X.~Yuan, X.~Zheng, Y.~Li, J.~Zhang, C.~Zhang, M.~Chen, J.~Liu, Z.~Fang, W.~Wang, J.~Xue, Y.~Tao, J.~Zhu, K.~Liu, S.~Lin, Y.~Sun, Y.~Li, D.~Wang, M.~Chen, Z.~Hu, X.~Xiao, Y.~Chen, Y.~Liu, W.~Liu, D.~Wang, Y.~Yang, J.~Jiang, and Q.~Lu.
\newblock {Hunyuan-DiT: A Powerful Multi-Resolution Diffusion Transformer with Fine-Grained Chinese Understanding}, 2024{\natexlab{b}}.
\newblock URL \url{https://arxiv.org/abs/2405.08748}.

\bibitem[Lin et~al.(2014)Lin, Maire, Belongie, Hays, Perona, Ramanan, Doll{\'{a}}r, and Zitnick]{coco}
T.~Lin, M.~Maire, S.~J. Belongie, J.~Hays, P.~Perona, D.~Ramanan, P.~Doll{\'{a}}r, and C.~L. Zitnick.
\newblock {Microsoft {COCO:} Common Objects in Context}.
\newblock In \emph{Proceedings of the European Conference on Computer Vision}, volume 8693 of \emph{Lecture Notes in Computer Science}, pages 740--755. Springer, 2014.

\bibitem[Lin et~al.(2024)Lin, Pathak, Li, Li, Xia, Neubig, Zhang, and Ramanan]{vqascore}
Z.~Lin, D.~Pathak, B.~Li, J.~Li, X.~Xia, G.~Neubig, P.~Zhang, and D.~Ramanan.
\newblock {Evaluating Text-to-Visual Generation with Image-to-Text Generation}.
\newblock In \emph{Proceedings of the European Conference on Computer Vision}, volume 15067 of \emph{Lecture Notes in Computer Science}, pages 366--384. Springer, 2024.

\bibitem[Lipman et~al.(2023)Lipman, Chen, Ben{-}Hamu, Nickel, and Le]{flowmatching}
Y.~Lipman, R.~T.~Q. Chen, H.~Ben{-}Hamu, M.~Nickel, and M.~Le.
\newblock Flow matching for generative modeling.
\newblock In \emph{Proceedings of the International Conference on Learning Representations}. OpenReview.net, 2023.

\bibitem[Liu et~al.(2024{\natexlab{a}})Liu, Akhgari, Visheratin, Kamko, Xu, Shrirao, Lambert, Souza, Doshi, and Li]{playgroundv3}
B.~Liu, E.~Akhgari, A.~Visheratin, A.~Kamko, L.~Xu, S.~Shrirao, C.~Lambert, J.~Souza, S.~Doshi, and D.~Li.
\newblock {Playground v3: Improving Text-to-Image Alignment with Deep-Fusion Large Language Models}, 2024{\natexlab{a}}.
\newblock URL \url{https://arxiv.org/abs/2409.10695}.

\bibitem[Liu et~al.(2024{\natexlab{b}})Liu, Zeng, Ren, Li, Zhang, Yang, Jiang, Li, Yang, Su, Zhu, and Zhang]{groundingdino}
S.~Liu, Z.~Zeng, T.~Ren, F.~Li, H.~Zhang, J.~Yang, Q.~Jiang, C.~Li, J.~Yang, H.~Su, J.~Zhu, and L.~Zhang.
\newblock {Grounding {DINO:} Marrying {DINO} with Grounded Pre-training for Open-Set Object Detection}.
\newblock In \emph{Proceedings of the European Conference on Computer Vision}, volume 15105 of \emph{Lecture Notes in Computer Science}, pages 38--55. Springer, 2024{\natexlab{b}}.

\bibitem[Lv et~al.(2024)Lv, Wei, Zuo, and Wong]{place}
Z.~Lv, Y.~Wei, W.~Zuo, and K.~K. Wong.
\newblock {{PLACE:} Adaptive Layout-Semantic Fusion for Semantic Image Synthesis}.
\newblock In \emph{Proceedings of the IEEE/CVF Conference on Computer Vision and Pattern Recognition}, pages 9264--9274. {IEEE}, 2024.

\bibitem[Ohanyan et~al.(2024)Ohanyan, Manukyan, Wang, Navasardyan, and Shi]{Zeropainter}
M.~Ohanyan, H.~Manukyan, Z.~Wang, S.~Navasardyan, and H.~Shi.
\newblock {Zero-Painter: Training-Free Layout Control for Text-to-Image Synthesis}.
\newblock In \emph{Proceedings of the IEEE/CVF Conference on Computer Vision and Pattern Recognition}, pages 8764--8774. {IEEE}, 2024.

\bibitem[openfree(2025)]{ghibli}
openfree.
\newblock {flux-chatgpt-ghibli-lora}.
\newblock \url{https://huggingface.co/openfree/flux-chatgpt-ghibli-lora}, 2025.

\bibitem[PatrickStarrrr(2024)]{oilpainting}
PatrickStarrrr.
\newblock {FLUX - Oil painting}.
\newblock \url{https://civitai.com/models/1455014/chatgpt-4o-renderer?modelVersionId=1697982}, 2024.

\bibitem[Peebles and Xie(2023)]{dit}
W.~Peebles and S.~Xie.
\newblock {Scalable Diffusion Models with Transformers}.
\newblock In \emph{Proceedings of the IEEE/CVF International Conference on Computer Vision}, pages 4172--4182. {IEEE}, 2023.

\bibitem[Phung et~al.(2024)Phung, Ge, and Huang]{attnrefocus}
Q.~Phung, S.~Ge, and J.~Huang.
\newblock {Grounded Text-to-Image Synthesis with Attention Refocusing}.
\newblock In \emph{Proceedings of the IEEE/CVF Conference on Computer Vision and Pattern Recognition}, pages 7932--7942. {IEEE}, 2024.

\bibitem[Podell et~al.(2024)Podell, English, Lacey, Blattmann, Dockhorn, M{\"{u}}ller, Penna, and Rombach]{sdxl}
D.~Podell, Z.~English, K.~Lacey, A.~Blattmann, T.~Dockhorn, J.~M{\"{u}}ller, J.~Penna, and R.~Rombach.
\newblock {{SDXL:} Improving Latent Diffusion Models for High-Resolution Image Synthesis}.
\newblock In \emph{Proceedings of the International Conference on Learning Representations}. OpenReview.net, 2024.

\bibitem[{Qwen Team}(2025)]{qwen2.5-VL}
{Qwen Team}.
\newblock {Qwen2.5-VL}, January 2025.
\newblock URL \url{https://qwenlm.github.io/blog/qwen2.5-vl/}.

\bibitem[Ramesh et~al.(2022)Ramesh, Dhariwal, Nichol, Chu, and Chen]{dalle2}
A.~Ramesh, P.~Dhariwal, A.~Nichol, C.~Chu, and M.~Chen.
\newblock {Hierarchical Text-Conditional Image Generation with CLIP Latents}, 2022.
\newblock URL \url{https://arxiv.org/abs/2204.06125}.

\bibitem[Rombach et~al.(2022)Rombach, Blattmann, Lorenz, Esser, and Ommer]{ldm}
R.~Rombach, A.~Blattmann, D.~Lorenz, P.~Esser, and B.~Ommer.
\newblock {High-Resolution Image Synthesis with Latent Diffusion Models}.
\newblock In \emph{Proceedings of the IEEE/CVF Conference on Computer Vision and Pattern Recognition}, pages 10674--10685. {IEEE}, 2022.

\bibitem[Ronneberger et~al.(2015)Ronneberger, Fischer, and Brox]{unet}
O.~Ronneberger, P.~Fischer, and T.~Brox.
\newblock {U-Net: Convolutional Networks for Biomedical Image Segmentation}.
\newblock In \emph{Medical Image Computing and Computer-Assisted Intervention}, volume 9351 of \emph{Lecture Notes in Computer Science}, pages 234--241. Springer, 2015.

\bibitem[Saharia et~al.(2022)Saharia, Chan, Saxena, Li, Whang, Denton, Ghasemipour, Lopes, Ayan, Salimans, Ho, Fleet, and Norouzi]{Imagen}
C.~Saharia, W.~Chan, S.~Saxena, L.~Li, J.~Whang, E.~L. Denton, S.~K.~S. Ghasemipour, R.~G. Lopes, B.~K. Ayan, T.~Salimans, J.~Ho, D.~J. Fleet, and M.~Norouzi.
\newblock {Photorealistic Text-to-Image Diffusion Models with Deep Language Understanding}.
\newblock In \emph{Advances in Neural Information Processing Systems}, 2022.

\bibitem[Shirakawa and Uchida(2024)]{noisecollage}
T.~Shirakawa and S.~Uchida.
\newblock {NoiseCollage: {A} Layout-Aware Text-to-Image Diffusion Model Based on Noise Cropping and Merging}.
\newblock In \emph{Proceedings of the IEEE/CVF Conference on Computer Vision and Pattern Recognition}, pages 8921--8930. {IEEE}, 2024.

\bibitem[{stability.ai}(2024)]{sd3.5}
{stability.ai}.
\newblock {Stable Diffusion 3.5}.
\newblock \url{https://stability.ai/news/introducing-stable-diffusion-3-5}, Nov 2024.

\bibitem[Taghipour et~al.(2024)Taghipour, Ghahremani, Bennamoun, Rekavandi, Laga, and Boussaid]{boxbind}
A.~Taghipour, M.~Ghahremani, M.~Bennamoun, A.~M. Rekavandi, H.~Laga, and F.~Boussaid.
\newblock {Box It to Bind It: Unified Layout Control and Attribute Binding in T2I Diffusion Models}, 2024.
\newblock URL \url{https://arxiv.org/abs/2402.17910}.

\bibitem[Tancik et~al.(2020)Tancik, Srinivasan, Mildenhall, Fridovich{-}Keil, Raghavan, Singhal, Ramamoorthi, Barron, and Ng]{fourier}
M.~Tancik, P.~P. Srinivasan, B.~Mildenhall, S.~Fridovich{-}Keil, N.~Raghavan, U.~Singhal, R.~Ramamoorthi, J.~T. Barron, and R.~Ng.
\newblock {Fourier Features Let Networks Learn High Frequency Functions in Low Dimensional Domains}.
\newblock In \emph{Advances in Neural Information Processing Systems}, 2020.

\bibitem[vjleoliu(2025)]{cute3d}
vjleoliu.
\newblock {ChatGPT-4o Renderer}.
\newblock \url{https://civitai.com/models/1455014/chatgpt-4o-renderer?modelVersionId=1697982}, 2025.

\bibitem[Wang et~al.(2024)Wang, Darrell, Rambhatla, Girdhar, and Misra]{instancediffusion}
X.~Wang, T.~Darrell, S.~S. Rambhatla, R.~Girdhar, and I.~Misra.
\newblock {InstanceDiffusion: Instance-Level Control for Image Generation}.
\newblock In \emph{Proceedings of the IEEE/CVF Conference on Computer Vision and Pattern Recognition}, pages 6232--6242. {IEEE}, 2024.

\bibitem[Wu et~al.(2024)Wu, Zhou, Ma, Su, Ma, and Wang]{ifadapter}
Y.~Wu, X.~Zhou, B.~Ma, X.~Su, K.~Ma, and X.~Wang.
\newblock {IFAdapter: Instance Feature Control for Grounded Text-to-Image Generation}, 2024.
\newblock URL \url{https://arxiv.org/abs/2409.08240}.

\bibitem[Xie et~al.(2023)Xie, Li, Huang, Liu, Zhang, Zheng, and Shou]{boxdiff}
J.~Xie, Y.~Li, Y.~Huang, H.~Liu, W.~Zhang, Y.~Zheng, and M.~Z. Shou.
\newblock {BoxDiff: Text-to-Image Synthesis with Training-Free Box-Constrained Diffusion}.
\newblock In \emph{Proceedings of the IEEE/CVF International Conference on Computer Vision}, pages 7418--7427. {IEEE}, 2023.

\bibitem[Xu et~al.(2019)Xu, Sun, Zhang, Zhao, and Lin]{adanorm}
J.~Xu, X.~Sun, Z.~Zhang, G.~Zhao, and J.~Lin.
\newblock {Understanding and Improving Layer Normalization}.
\newblock In \emph{Advances in Neural Information Processing Systems}, pages 4383--4393, 2019.

\bibitem[Xue et~al.(2023)Xue, Huang, Sun, Song, and Zhang]{xue2023freestyle}
H.~Xue, Z.~Huang, Q.~Sun, L.~Song, and W.~Zhang.
\newblock {Freestyle Layout-to-Image Synthesis}.
\newblock In \emph{Proceedings of the IEEE/CVF Conference on Computer Vision and Pattern Recognition}, pages 14256--14266. {IEEE}, 2023.

\bibitem[Yang et~al.(2023{\natexlab{a}})Yang, Luo, Chen, Wang, Liang, and Lin]{lawdiff}
B.~Yang, Y.~Luo, Z.~Chen, G.~Wang, X.~Liang, and L.~Lin.
\newblock {LAW-Diffusion: Complex Scene Generation by Diffusion with Layouts}.
\newblock In \emph{Proceedings of the IEEE/CVF International Conference on Computer Vision}, pages 22612--22622. {IEEE}, 2023{\natexlab{a}}.

\bibitem[Yang et~al.(2023{\natexlab{b}})Yang, Wang, Gan, Li, Lin, Wu, Duan, Liu, Liu, Zeng, and Wang]{reco}
Z.~Yang, J.~Wang, Z.~Gan, L.~Li, K.~Lin, C.~Wu, N.~Duan, Z.~Liu, C.~Liu, M.~Zeng, and L.~Wang.
\newblock {ReCo: Region-Controlled Text-to-Image Generation}.
\newblock In \emph{Proceedings of the IEEE/CVF Conference on Computer Vision and Pattern Recognition}, pages 14246--14255. {IEEE}, 2023{\natexlab{b}}.

\bibitem[Yao et~al.(2024)Yao, Yu, Zhang, Wang, Cui, Zhu, Cai, Li, Zhao, He, Chen, Zhou, Zou, Zhang, Hu, Zheng, Zhou, Cai, Han, Zeng, Li, Liu, and Sun]{minicpm}
Y.~Yao, T.~Yu, A.~Zhang, C.~Wang, J.~Cui, H.~Zhu, T.~Cai, H.~Li, W.~Zhao, Z.~He, Q.~Chen, H.~Zhou, Z.~Zou, H.~Zhang, S.~Hu, Z.~Zheng, J.~Zhou, J.~Cai, X.~Han, G.~Zeng, D.~Li, Z.~Liu, and M.~Sun.
\newblock {MiniCPM-V: A GPT-4V Level MLLM on Your Phone}, 2024.
\newblock URL \url{https://arxiv.org/abs/2408.01800}.

\bibitem[Zhang et~al.(2025)Zhang, Hong, Wang, Shao, Wu, Wu, and Jiang]{creatilayout}
H.~Zhang, D.~Hong, Y.~Wang, J.~Shao, X.~Wu, Z.~Wu, and Y.-G. Jiang.
\newblock {CreatiLayout: Siamese Multimodal Diffusion Transformer for Creative Layout-to-Image Generation}, 2025.
\newblock URL \url{https://arxiv.org/abs/2412.03859}.

\bibitem[Zhang et~al.(2023)Zhang, Rao, and Agrawala]{controlnet}
L.~Zhang, A.~Rao, and M.~Agrawala.
\newblock Adding conditional control to text-to-image diffusion models.
\newblock In \emph{Proceedings of the IEEE/CVF International Conference on Computer Vision}, pages 3813--3824. {IEEE}, 2023.

\bibitem[Zheng et~al.(2023)Zheng, Zhou, Li, Qi, Shan, and Li]{layoutdiffusion}
G.~Zheng, X.~Zhou, X.~Li, Z.~Qi, Y.~Shan, and X.~Li.
\newblock {LayoutDiffusion: Controllable Diffusion Model for Layout-to-Image Generation}.
\newblock In \emph{Proceedings of the IEEE/CVF Conference on Computer Vision and Pattern Recognition}, pages 22490--22499. {IEEE}, 2023.

\bibitem[Zhou et~al.(2024)Zhou, Li, Ma, Zhang, and Yang]{migc}
D.~Zhou, Y.~Li, F.~Ma, X.~Zhang, and Y.~Yang.
\newblock {{MIGC:} Multi-Instance Generation Controller for Text-to-Image Synthesis}.
\newblock In \emph{Proceedings of the IEEE/CVF Conference on Computer Vision and Pattern Recognition}, pages 6818--6828. {IEEE}, 2024.

\bibitem[Zhou et~al.(2025{\natexlab{a}})Zhou, Li, Ma, Yang, and Yang]{migc++}
D.~Zhou, Y.~Li, F.~Ma, Z.~Yang, and Y.~Yang.
\newblock {{MIGC++:} Advanced Multi-Instance Generation Controller for Image Synthesis}.
\newblock \emph{IEEE Transactions on Pattern Analysis and Machine Intelligence}, 47\penalty0 (3):\penalty0 1714--1728, 2025{\natexlab{a}}.

\bibitem[Zhou et~al.(2025{\natexlab{b}})Zhou, Xie, Yang, and Yang]{3disflux}
D.~Zhou, J.~Xie, Z.~Yang, and Y.~Yang.
\newblock {3DIS-FLUX: simple and efficient multi-instance generation with DiT rendering}, 2025{\natexlab{b}}.
\newblock URL \url{https://arxiv.org/abs/2501.05131}.

\end{thebibliography}


\newpage
\appendix

\section*{\centering \LARGE Supplementary Material}
\section{Early‐Stage Layout Control in Assemble‑MMDiT}

We apply the Assemble-MMDiT layout control module exclusively during the initial 30\% of the denoising trajectory and deactivate it for the remaining 70\%. This two-stage strategy provides robust low-frequency structural guidance in the early stages to facilitate layout alignment, while allowing subsequent unconstrained refinement of high-frequency details during later denoising phases.

\begin{figure}[htbp]
    \centering
    {\includegraphics[width=0.9\textwidth]{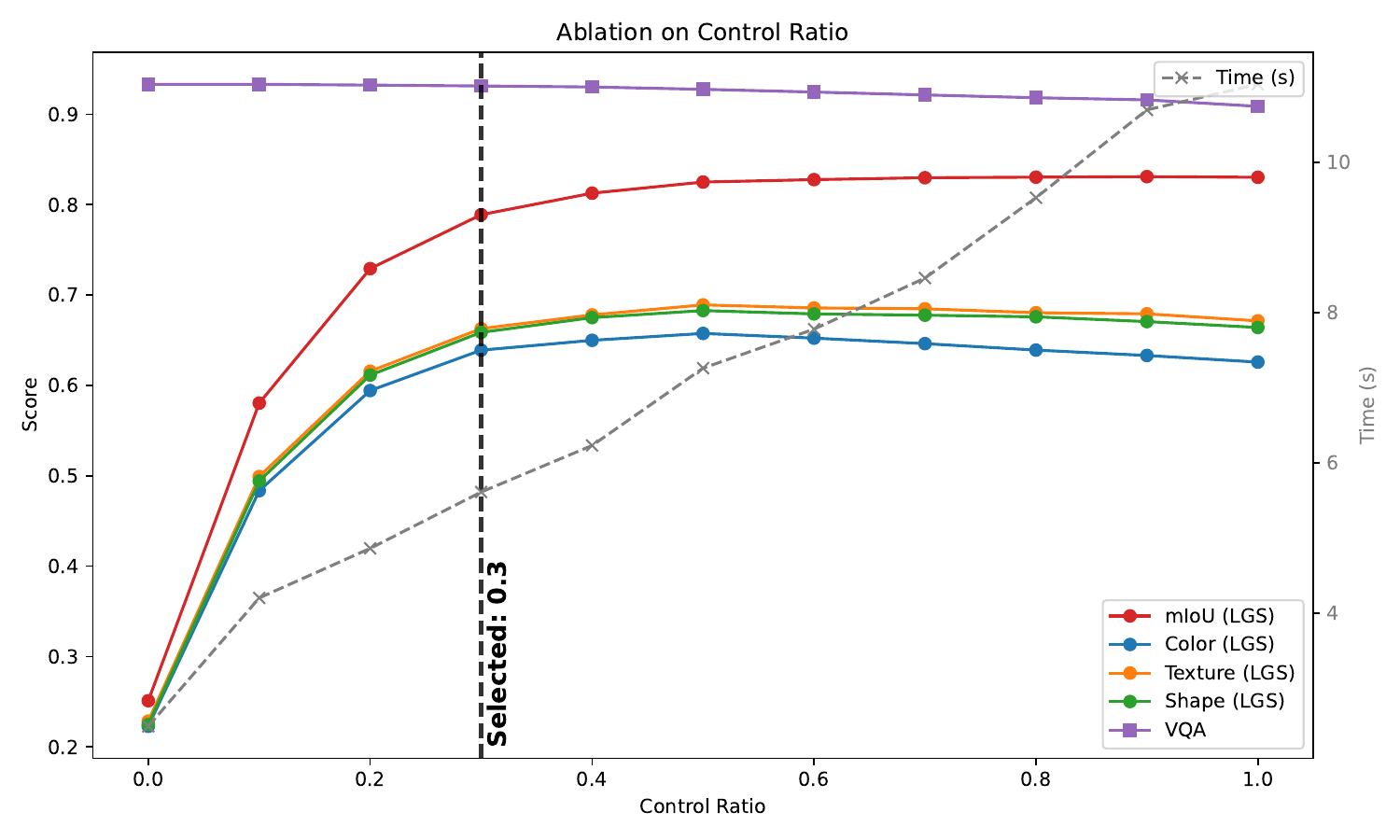}}
    \caption{Impact of the proportion of diffusion steps incorporating layout conditioning on generation quality.}
    \label{fig:ratio_layoutcontrol}
\end{figure}

As illustrated in Figure~\ref{fig:ratio_layoutcontrol}, restricting layout control to less than 30\% of the diffusion process results in insufficient layout alignment with the target bounding boxes. In contrast, extending control beyond this optimal threshold leads to a decline in output quality. Furthermore, increasing the proportion of layout-guided steps results in significant additional computational cost.

\section{Additional Ablation Studies}

\subsection{Effect of Bbox Encoding and DenseSample}
To clarify the role of bounding box encoding and DenseSample, we further ablated the SD3-M based InstanceAssemble model on DenseLayout. 
Bounding box embeddings guide correct object placement, while DenseSample provides additional improvements in spatial accuracy and instance-level semantics. 
The results in Table~\ref{tab:appendix_bbox} demonstrate that both components contribute to the overall performance.
\begin{table}[h]
\centering
\scriptsize
\caption{Ablation on bounding box encoding and DenseSample.}
\label{tab:appendix_bbox}
\begin{tabular}{lccccc}
\toprule
Setting & mIoU$\uparrow$ & color$\uparrow$ & texture$\uparrow$ & shape$\uparrow$ & VQA$\uparrow$ \\
\midrule
w/o bbox encoding, w/o DenseSample & 51.22 & 32.15 & 34.04 & 33.53 & 93.30 \\
w/ bbox encoding, w/o DenseSample  & 51.28 & 32.68 & 34.94 & 34.58 & 93.33 \\
w/ bbox encoding, w/ DenseSample   & \textbf{52.07} & \textbf{33.77} & \textbf{36.21} & \textbf{35.81} & \textbf{93.54} \\
\bottomrule
\end{tabular}
\end{table}

\subsection{Comparison with Attention Mask-based Region Injection}
We also compare our Assemble-Attn design with attention mask-based region injection. 
While both can be viewed as region-wise attention mechanisms, attention masks operate globally and may cause semantic leakage in overlapping regions. 
Our method instead applies instance-wise self-attention on cropped latent regions and then fuses the updated features via the Assemble step, which is more effective in dense layouts. 
As shown in Table~\ref{tab:appendix_mask}, our design achieves superior instance attribute consistency and a higher VQA score compared to the attention mask baseline.
\begin{table}[h]
\centering
\scriptsize
\caption{Comparison between attention mask-based injection and our Assemble-Attn.}
\label{tab:appendix_mask}
\begin{tabular}{lccccc}
\toprule
Method & spatial$\uparrow$ & color$\uparrow$ & texture$\uparrow$ & shape$\uparrow$ & VQA$\uparrow$ \\
\midrule
SD3-Medium (base model) & 77.49 & 60.28 & 62.55 & 60.38 & 93.30 \\
Attention mask (SD3-M)  & 94.11 & 74.28 & 77.58 & 76.54 & 91.53 \\
InstanceAssemble (ours, SD3-M) & \textbf{94.97} & \textbf{77.53} & \textbf{80.72} & \textbf{80.11} & \textbf{93.12} \\
\bottomrule
\end{tabular}
\end{table}

\section{Underlying data for radar-chart visualizations}

In Sections~\ref{sec:eval_addvis}, we utilized a radar chart to depict each quantitative variable along equi-angular axes, providing an intuitive comparison. This visualization highlights the multifaceted superiority of our method. Here, we present the corresponding raw evaluation results in tabular form. Specifically, \Tabref{tab:densevisual} corresponds to \Figref{fig:visualradar}, thus ensuring a clear mapping between each radar-chart subfigure and its underlying data.

\tabdensevisual

\section{More Details on DenseLayout Evaluation Dataset}
\subsection{Construction Pipeline of DenseLayout Dataset}

\begin{figure}[htbp]
    \centering
    \includegraphics[width=0.6\textwidth]{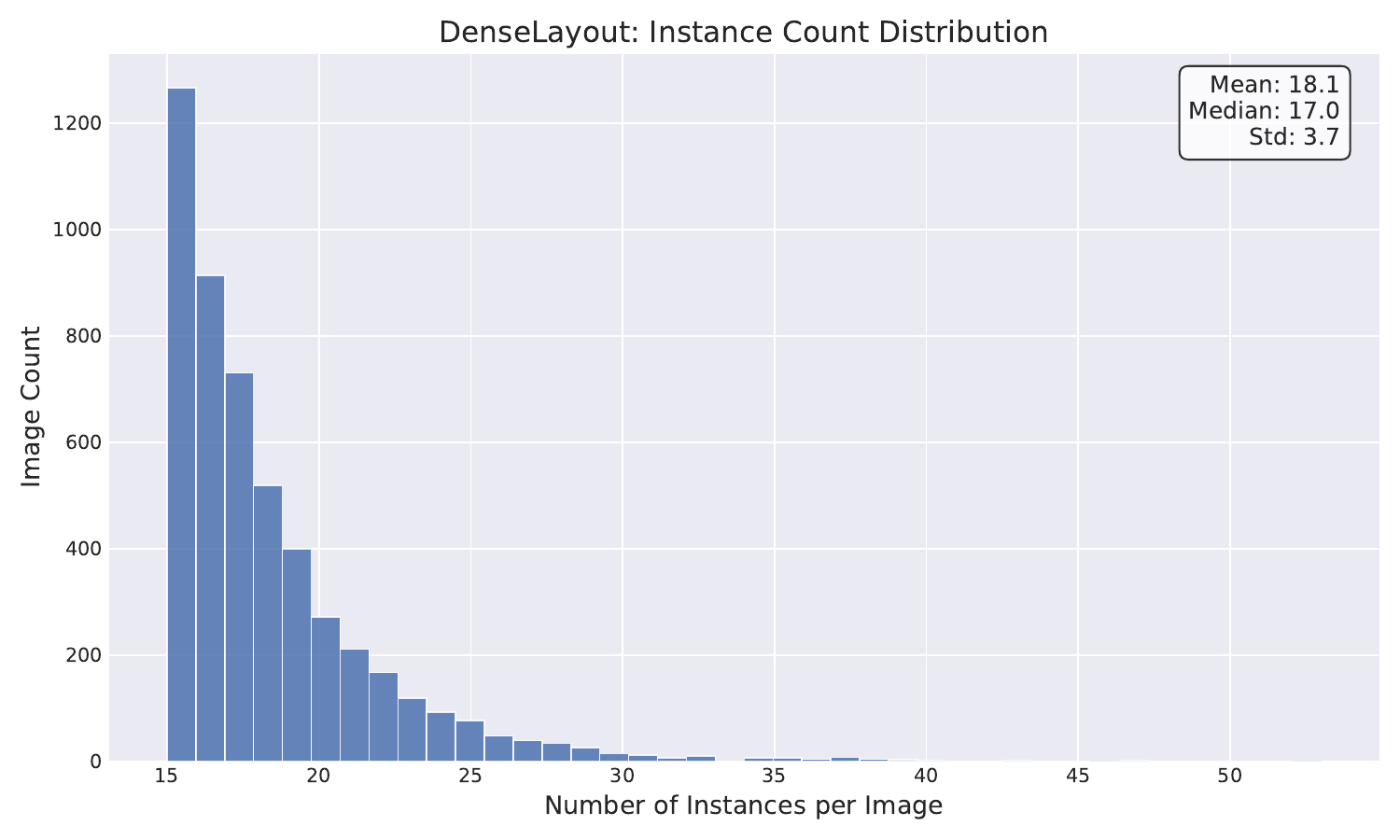}
    \caption{Instance count distribution per image in DenseLayout.}
\label{fig:instance_dist}
\end{figure}

The DenseLayout dataset is constructed through a multi-stage pipeline designed to extract high-density and semantically-rich layout information from synthetic images. 
The pipeline includes following steps:

\begin{enumerate}
    \item \textbf{Image Generation using Flux.1-Dev~\cite{flux}} \\
    A diverse set of synthetic images is generated using Flux.1-Dev, a text-to-image model. The input prompts are generic textual descriptions, sampled from the LayoutSAM dataset, which is based on SA-1B. The images are resized to maintain the same aspect ratio as the original SA-1B images, with the longer edge set to 1024 pixels. This step provides a visually complex base for extracting layout structures.

    \item \textbf{Multi-label Tagging using RAM++~\cite{ramplus}} \\
    The generated images are tagged using RAM++, the next-generation model of RAM, which supports open-set recognition. These tags offer high-level semantic guidance for subsequent grounding.

    \item \textbf{Object Detection via GroundingDINO~\cite{groundingdino}} \\
    Using the image and its predicted tags as input, GroundingDINO performs open-set object detection. It outputs bounding boxes and class labels for all detected entities. The detection is configured with a \texttt{box\_threshold} of 0.35 and a \texttt{text\_threshold} of 0.25. Each detected bounding box is treated as an \textbf{instance bounding box}, and the corresponding predicted label is recorded as the \textbf{instance description}.

    \item \textbf{Detailed Captioning with Qwen2.5-VL~\cite{qwen2.5-VL}} \\
    Each bounding box region is cropped from the original image and fed into Qwen2.5-VL to generate a fine-grained caption. These region-level captions are stored as the \textbf{detailed description} for each instance, enriching the semantic information beyond category labels.

    \item \textbf{Density Filtering} \\
    To ensure high layout complexity, only images with 15 or more detected instances (as output by GroundingDINO) are retained. This results in a dense layout distribution suitable for layout-conditioned generation tasks. The distribution of instance count is shown in \figref{fig:instance_dist}.
\end{enumerate}
Finally, the DenseLayout dataset contains \textbf{5,000 images} and \textbf{90,339 instances}, with an average of \textbf{18.1 instances per image}.

\paragraph{Annotation Format.}
The annotation for each image consists of:

\begin{itemize}
    \item \texttt{global\_caption}: the original prompt used for image generation.
    \item \texttt{image\_info}: metadata of the image, including height and width.
    \item \texttt{instance\_info}: a list of instances, each with:
    \begin{itemize}
        \item \texttt{bbox}: the bounding box of the instance, formatting as $[x_1, y_1, x_2, y_2]$.
        \item \texttt{description}: the category label predicted by GroundingDINO.
        \item \texttt{detail\_description}: a fine-grained caption generated by Qwen2.5-VL for the cropped region.
    \end{itemize}
\end{itemize}

\subsection{Samples of DenseLayout Dataset}

\lstdefinestyle{jsonstyle}{
    basicstyle=\ttfamily\tiny,
    backgroundcolor=\color{gray!10},
    frame=single,
    breaklines=true,
    showstringspaces=false
}

\begin{figure*}[htbp]
    \vspace{20pt}
    \centering
    \begin{subfigure}[t]{0.4\textwidth}
        \centering
        {\includegraphics[clip, trim=0.5cm 11cm 0.5cm 5cm, width=\linewidth]{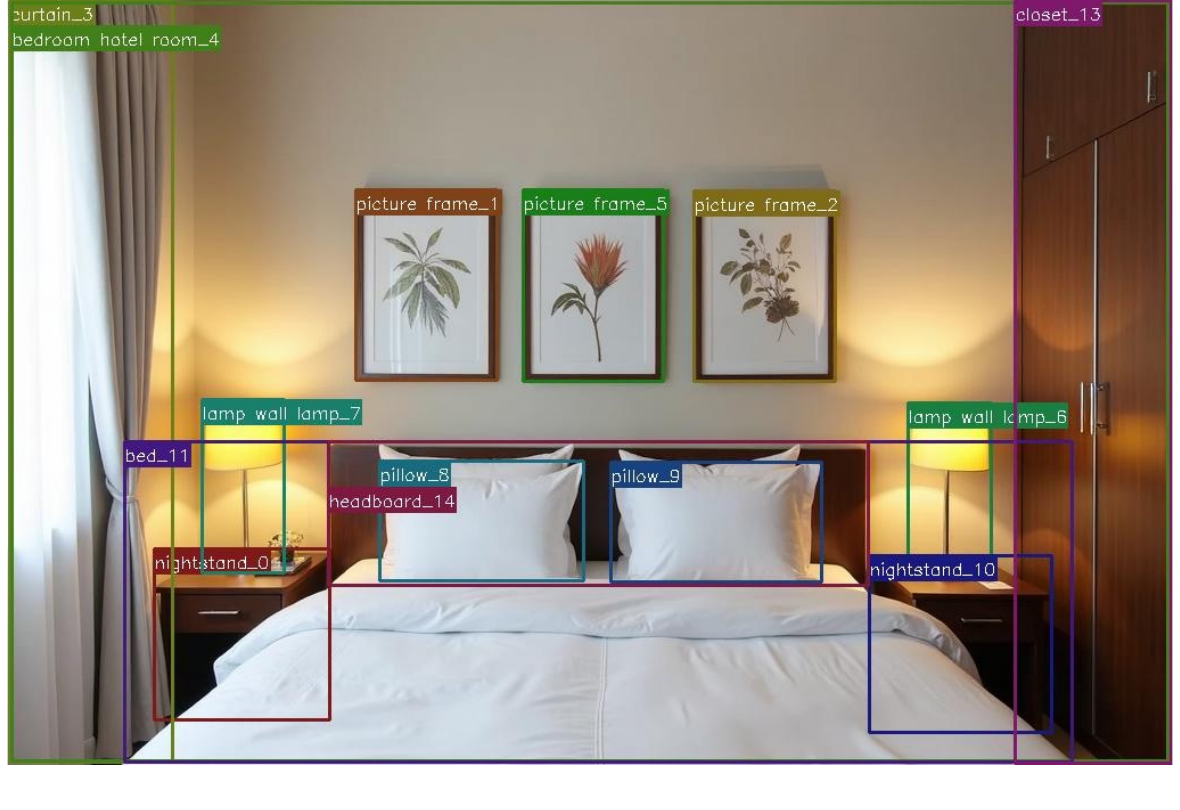}}
    \end{subfigure}
    \hfill
    \begin{subfigure}[t]{0.58\textwidth}
        \centering
        \lstset{style=jsonstyle}
        \begin{minipage}{\linewidth}
        \vspace{-80pt}
        \begin{lstlisting}
"instance_info": [
{
    "bbox": [129,489,283,642],
    "description": "nightstand",
    "detail_description": "The nightstand is dark brown, compact, with a drawer."
},
{
    "bbox": [306,170,430,339],
    "description": "picture frame",
    "detail_description": "Brown wooden frame containing a watercolor painting of green leaves on a white background."
},
{
    "bbox": [603,170,727,340],
    "description": "picture frame",
    "detail_description": "A simple brown wooden frame holds a botanical print with detailed leaves and stems."
},
...
]
        \end{lstlisting}
        \end{minipage}
    \end{subfigure}

    \caption{A sample of DenseLayout and its annotation.}
    \label{fig:onesample}
\end{figure*}

\begin{figure}[htbp]
    \centering
    {\includegraphics[clip, trim=0cm 20cm 15cm 0cm, width=\textwidth]{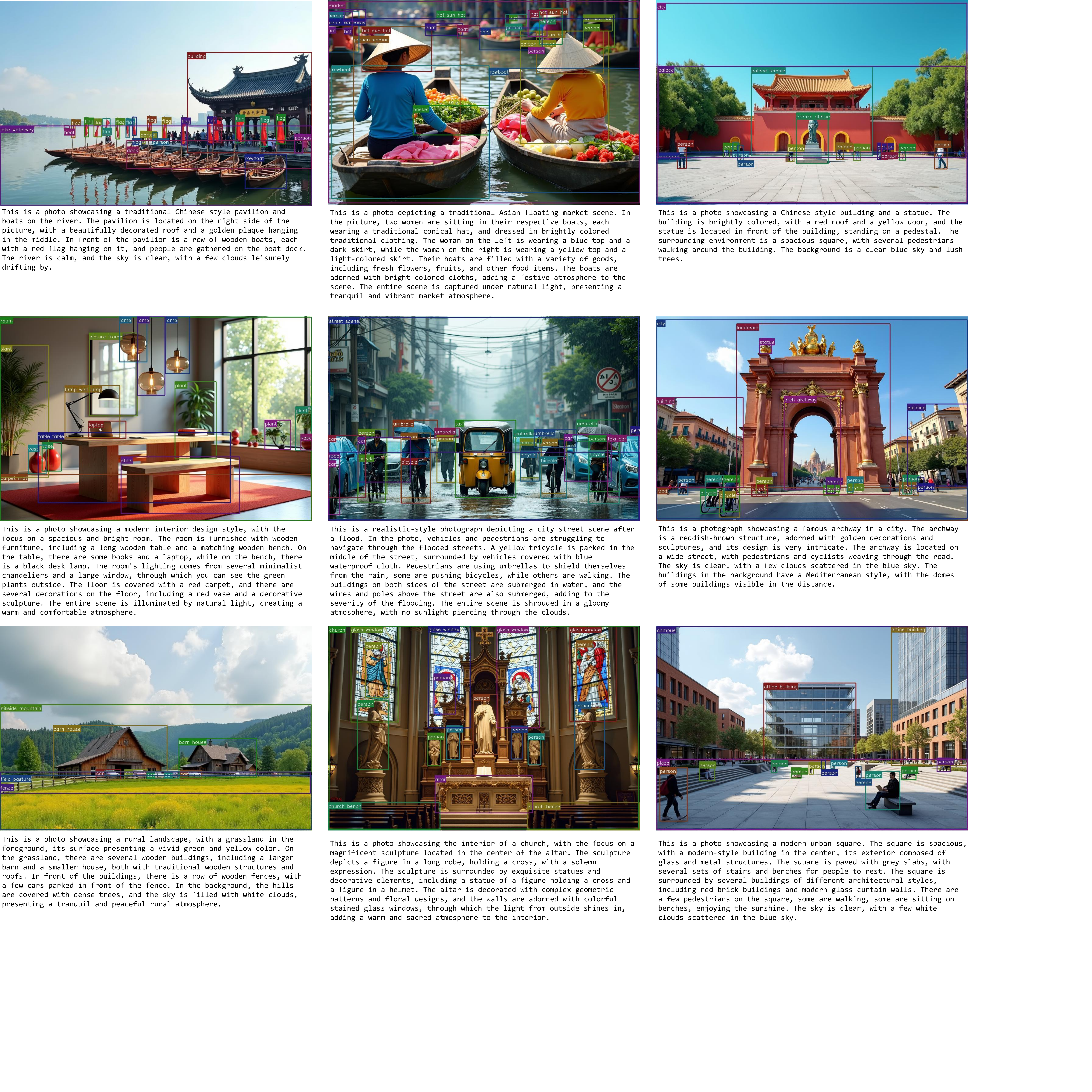}}
    \caption{Representative samples from DenseLayout dataset demonstrating: (a) High-density scene with $\geq15$ instances, (b) Complex instance relationships with precise attribute specifications.}
\label{fig:dense_layout_samples}
\end{figure}

\clearpage
\section{More results with textual-only content}
\label{suppl:quali}

\subsection{InstanceAssemble based on SD3-Medium}
\begin{figure}[h]
    \centering
    {\includegraphics[clip, trim=0cm 93cm 15cm 0cm, width=0.98\textwidth]{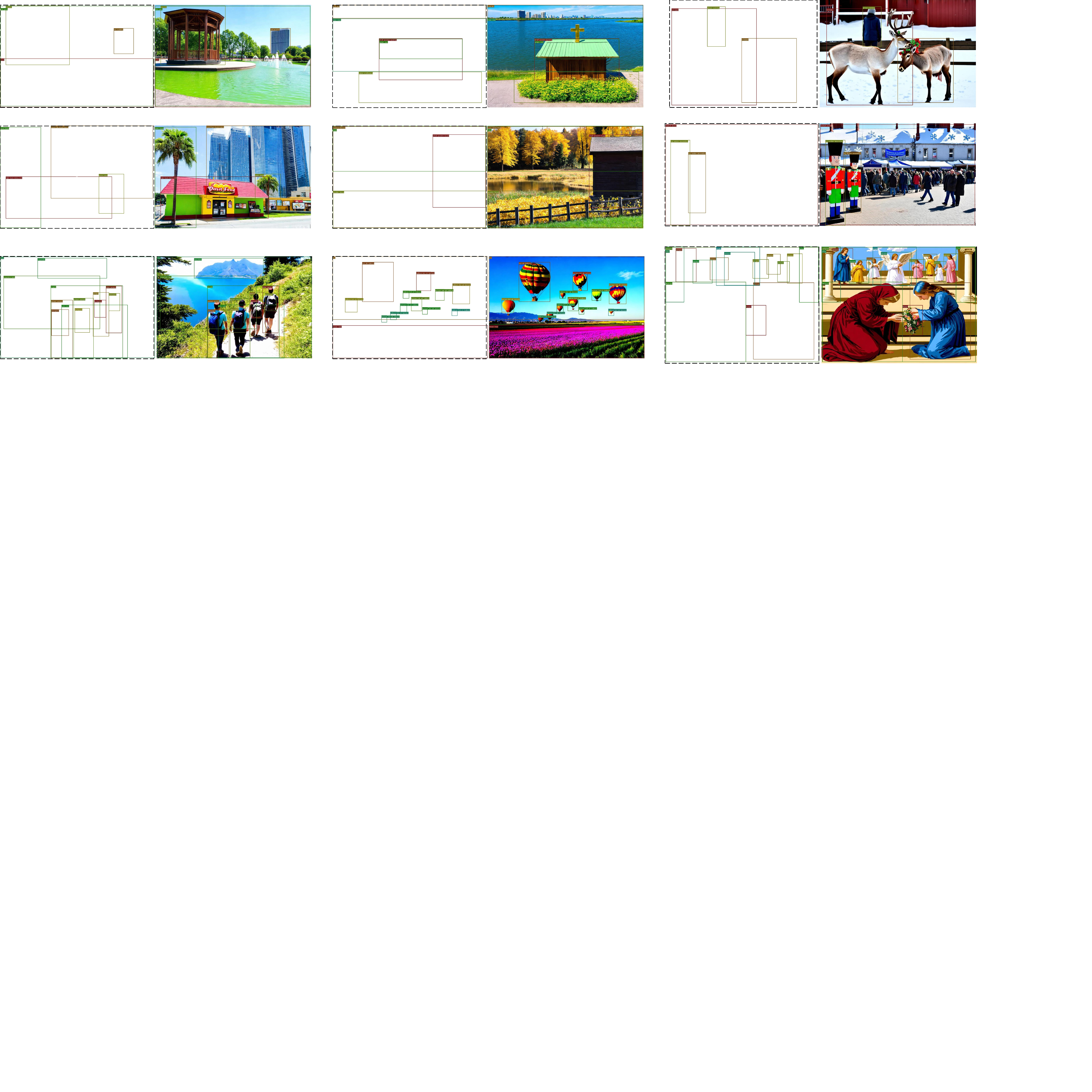}}
    \caption{\footnotesize More results of InstanceAssemble based on SD3-Medium with textual-only content.}
\label{fig:suppl_sd3}
\end{figure}

\subsection{InstanceAssemble based on Flux.1-Dev}
\begin{figure}[h]
    \centering
    {\includegraphics[clip, trim=0cm 94cm 12cm 0cm, width=0.98\textwidth]{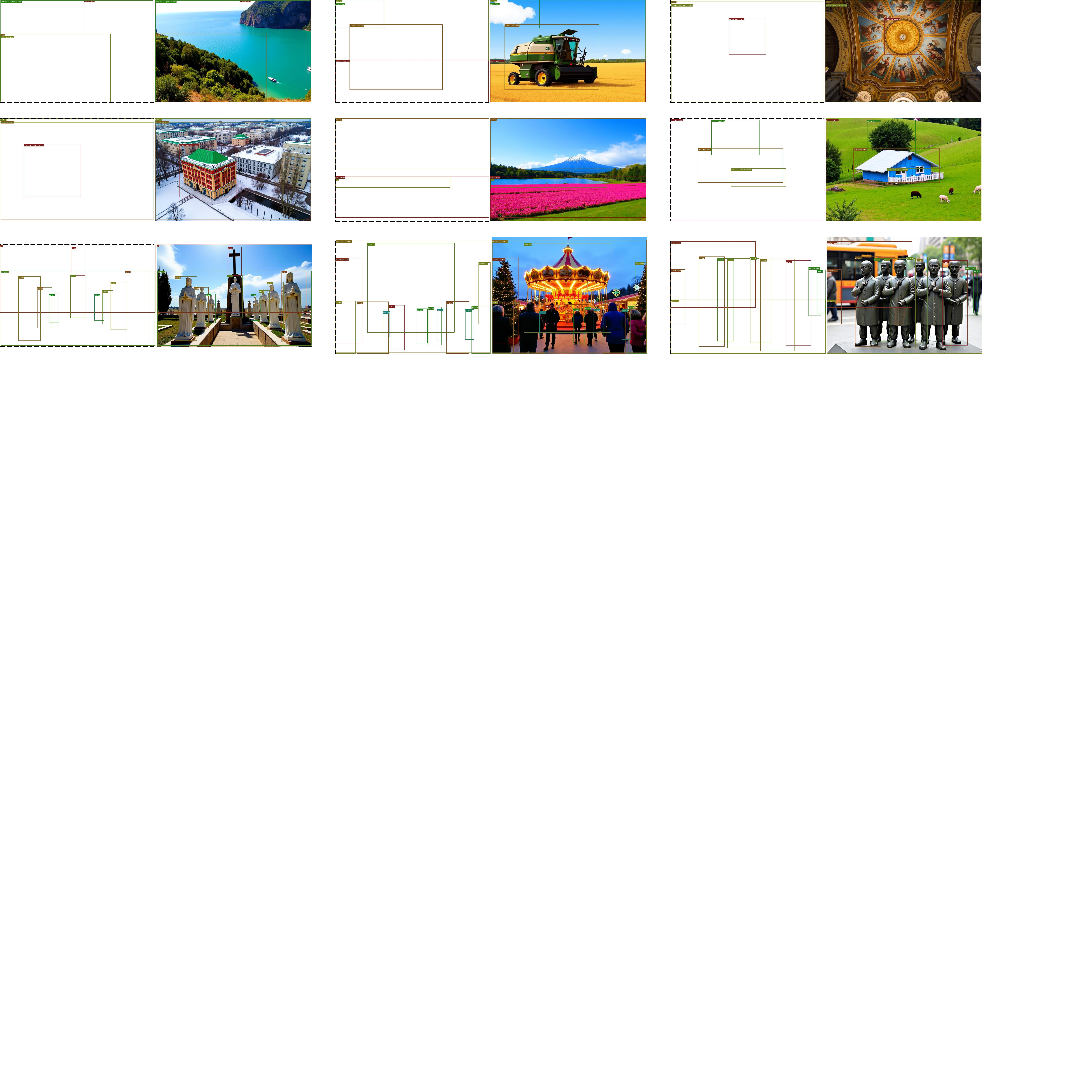}}
    \caption{\footnotesize More results of InstanceAssemble based on Flux.1-Dev with textual-only content.}
\label{fig:suppl_fluxdev}
\end{figure}

\subsection{InstanceAssemble based on Flux.1-Schnell}
\begin{figure}[h]
    \centering
    {\includegraphics[clip, trim=0cm 94cm 10cm 0cm, width=0.98\textwidth]{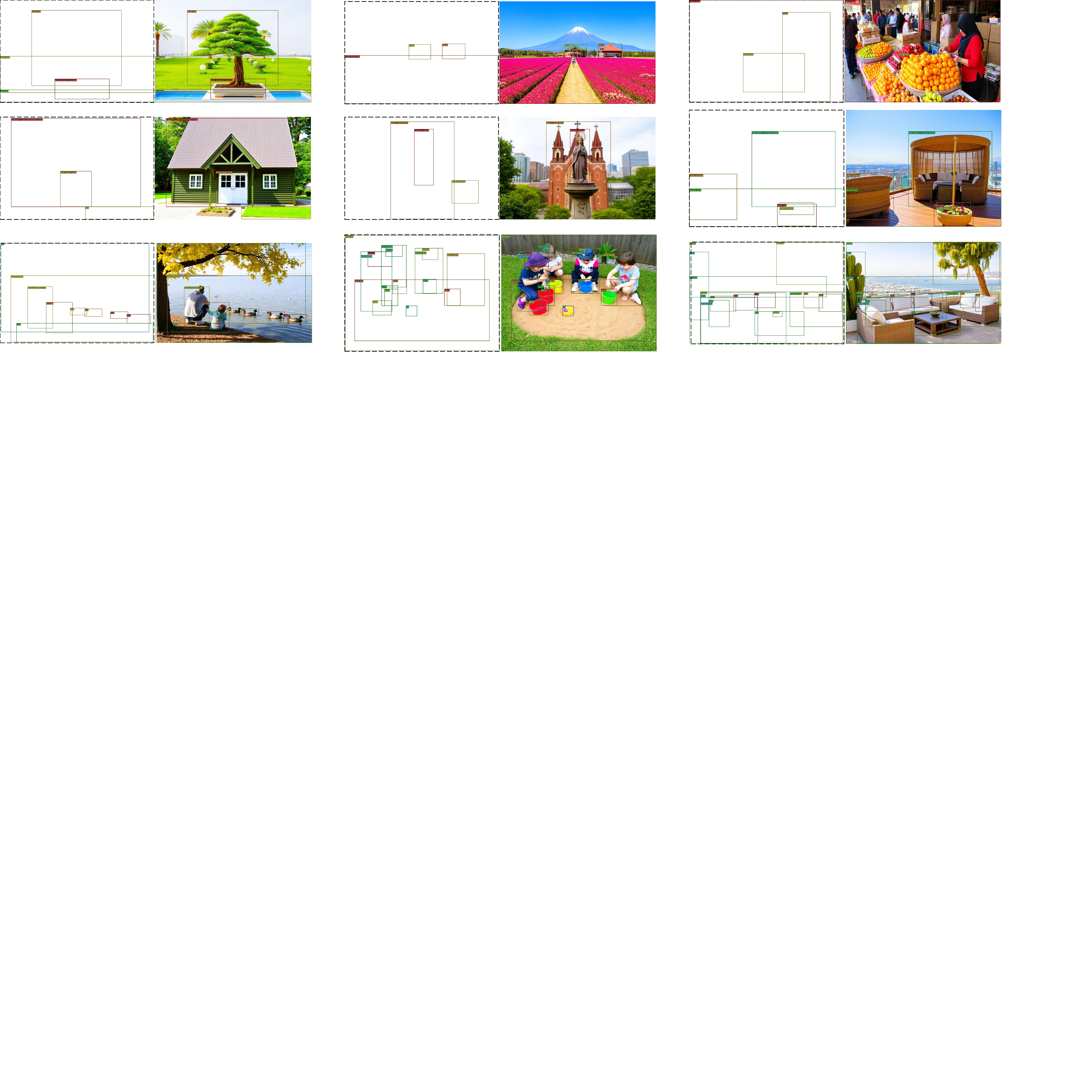}}
    \caption{\footnotesize More results of InstanceAssemble based on Flux.1-Schnell with textual-only content.}
\label{fig:suppl_fluxschnell}
\end{figure}

\clearpage
\section{More results with additional visual content}
\subsection{Additional Image}

\begin{figure}[htbp]
    \centering
    {\includegraphics[clip, trim=0cm 105cm 40cm 0cm, width=0.98\textwidth]{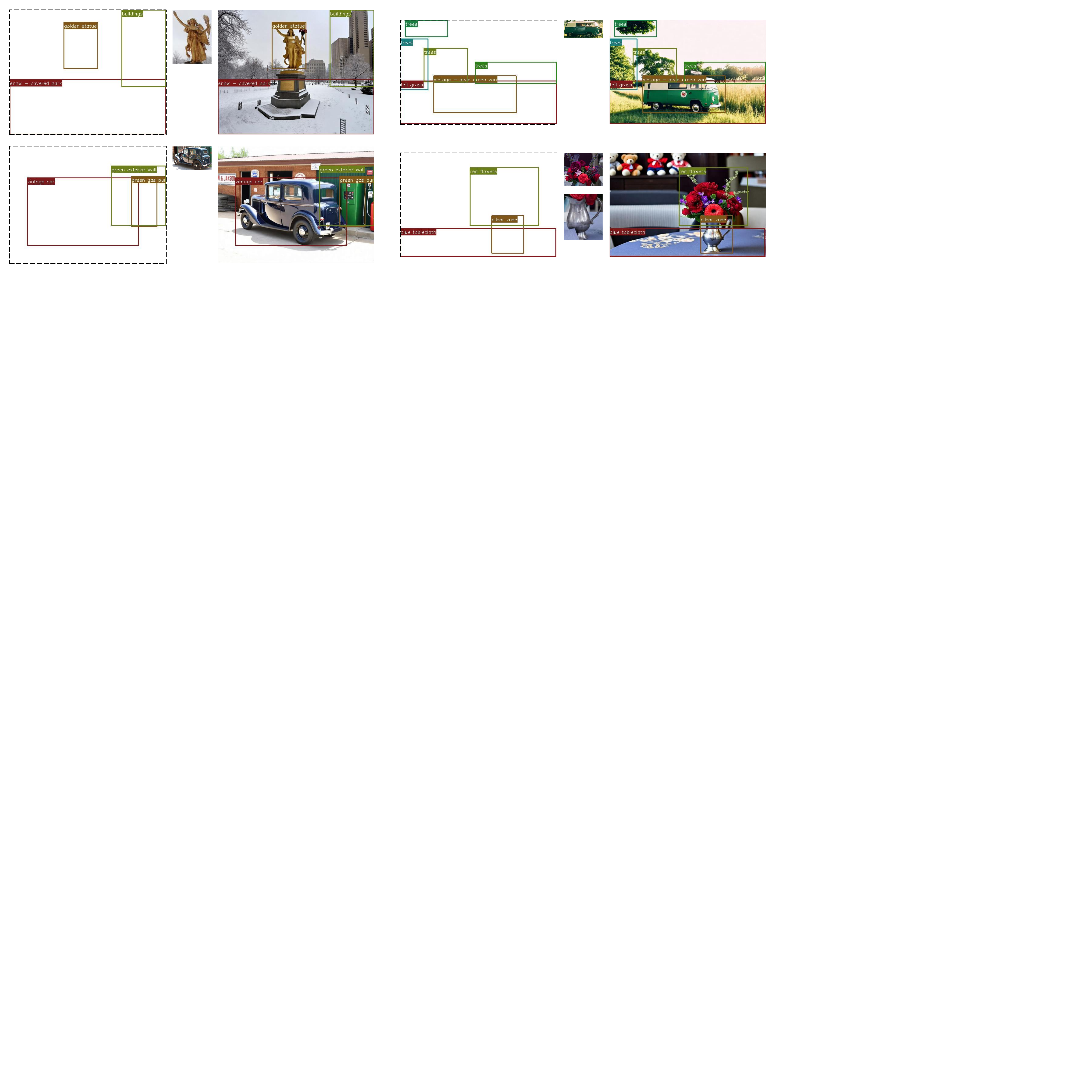}}
    \caption{More results with additional image.}
\end{figure}

\subsection{Additional Depth}
\begin{figure}[htbp]
    \centering
    {\includegraphics[clip, trim=0cm 108cm 40cm 0cm, width=0.98\textwidth]{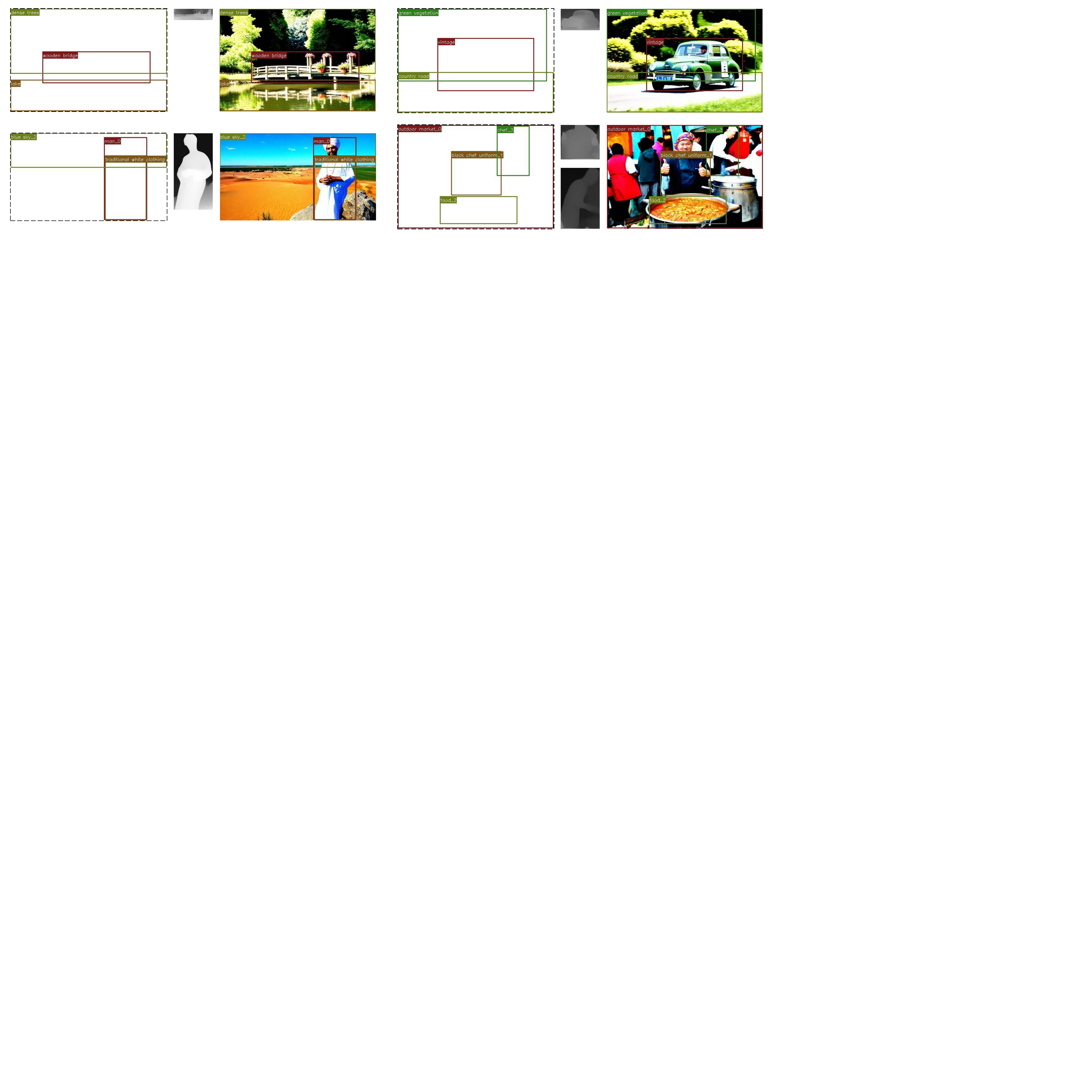}}
    \caption{More results with additional depth.}
\end{figure}

\subsection{Additional Edge}
\begin{figure}[htbp]
    \centering
    {\includegraphics[clip, trim=0cm 93cm 40cm 1cm, width=0.98\textwidth]{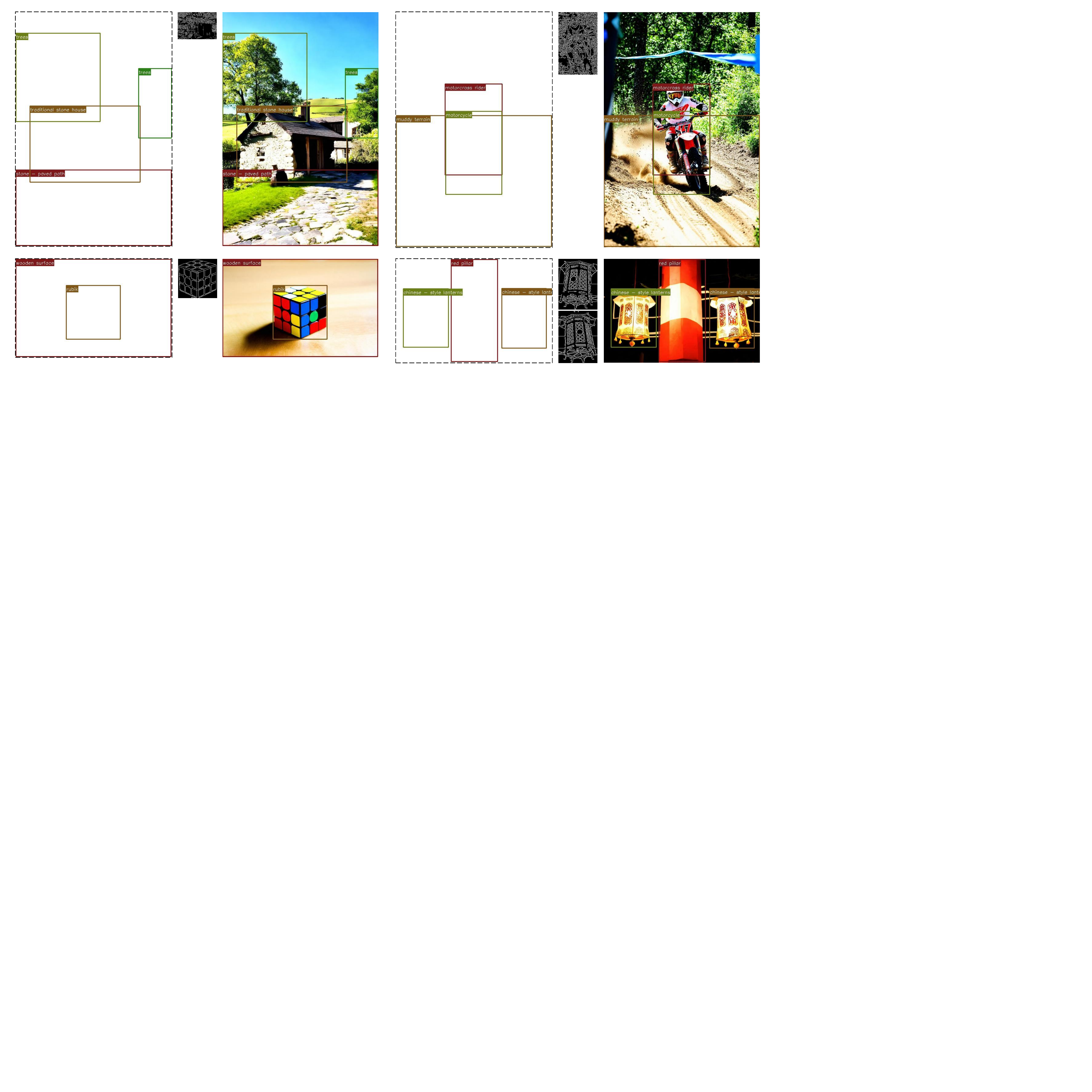}}
    \caption{More results with additional edge.}
\end{figure}

\end{document}